\documentclass[11pt]{article}

\usepackage[final]{acl}
\usepackage{dblfloatfix}

\usepackage{times}
\usepackage{latexsym}

\usepackage[T1]{fontenc}

\usepackage[utf8]{inputenc}
\usepackage{microtype}
\usepackage{inconsolata}
\usepackage{xcolor}
\usepackage{colortbl} 

\usepackage{graphicx}          
\usepackage{subcaption}        
\usepackage{caption}           
\usepackage{graphicx}
\usepackage{lipsum} 
\usepackage{multirow}
\usepackage{amsmath,amssymb}

\usepackage{tikz}
\usetikzlibrary{positioning,shapes.misc,shadows.blur,arrows.meta}
\usepackage{fontawesome5}

\usepackage{mathtools}

\usepackage{enumitem}

\usepackage{booktabs}
\usepackage{siunitx}

\usepackage{tabularx}
\usepackage{array}

\usepackage{url}

\definecolor{darkred}{HTML}{8B0000}   
\definecolor{darkblue}{HTML}{191970}  
\definecolor{cbblue}{RGB}{86,180,233}   
\definecolor{cborange}{RGB}{230,159,0}  

\usepackage{tcolorbox}
\usepackage{listings}
\tcbuselibrary{skins,breakable}

\definecolor{TitleRed}{HTML}{FF6B6B}
\definecolor{LightGray}{gray}{0.95} 
\definecolor{FrameGray}{gray}{0.70} 
\definecolor{TitleGray}{gray}{0.55} 

\lstdefinestyle{jsonstyle}{
  basicstyle=\ttfamily\small,
  columns=fullflexible,
  breaklines=true,
  frame=none,
  showstringspaces=false
}

\newtcolorbox{PromptCard}[1][]{%
  enhanced,
  breakable,
  colback=LightGray,
  colframe=FrameGray,
  boxrule=1.4pt,
  arc=8pt,
  left=10pt,right=10pt,top=10pt,bottom=10pt,
  title={Prompt Template},
  colbacktitle=TitleGray,
  coltitle=white,
  fonttitle=\bfseries\large,
  attach boxed title to top left={xshift=0pt,yshift=-2mm},
  boxed title style={
    sharp corners,
    arc=8pt,
    boxrule=0pt,
    left=10pt,right=10pt,top=6pt,bottom=6pt
  },
  #1
}

\title{Understanding Reliability in LLM-based Human Behavior Simulation}

\author{
  Pei Wang, \ Lei Wang, \ Yuanzi Li, \ Xu Chen\thanks{\; Corresponding author.} \\
  Gaoling School of Artificial Intelligence, Renmin University of China \\
  \texttt{\{wang\_pei, xu.chen\}@ruc.edu.cn} \\}

\begin{document}
\maketitle
\begin{abstract}
Large language models (LLMs) are increasingly used to simulate human survey responses and behavioral reactions, yet unreliable simulations can mislead social science conclusions. However, existing evaluations focus on end-to-end scores, leaving it unclear how different aspects of the simulation process interact to determine reliability. We propose ReliMap, which decomposes LLM-based human behavior simulation into three structured layers and evaluates reliability at both the individual level (R1) and population level (R2) across three configuration dimensions: model capacity, profile completeness, and population coverage. Through experiments across four simulation tasks and eleven LLMs, we find that all models exhibit substantial distributional bias without profile conditioning. Profile conditioning reduces this bias with diminishing returns. Larger models benefit more, and attribute informativeness matters more than quantity. Critically, R1 gains do not reliably transfer to R2—individual and population-level reliability can move in opposite directions. At the population layer, increasing coverage reduces variance but not systematic bias, with R2 stabilizing at around 50–100 individuals. These findings highlight that reliable simulation cannot be achieved by optimizing any single layer in isolation, but requires coordinated improvement across all three.\footnote{Our project page is available at \url{https://yupei-wang.github.io/understand_reliability/}.}

\end{abstract}

\section{Introduction}
Human experiments and surveys remain the gold standard for studying behavioral tendencies and social attitudes, yet their scale is fundamentally constrained by cost and participant availability~\cite{aher2023usinglargelanguagemodels, Argyle_2023, bisbee2024synthetic}. Large language models (LLMs) have emerged as a promising alternative, enabling rapid, low-cost simulation of human responses at scale. This has sparked growing interest in using LLMs as computational proxies for human participants \cite{dominguezolmedo2024questioningsurveyresponseslarge, hu2024generativelanguagemodelsexhibit, manning2024automatedsocialsciencelanguage, binz2025foundation, hu2025simbenchbenchmarkingabilitylarge}.

\begin{figure}[tbp]
    \centering
\includegraphics[width=0.45\textwidth]{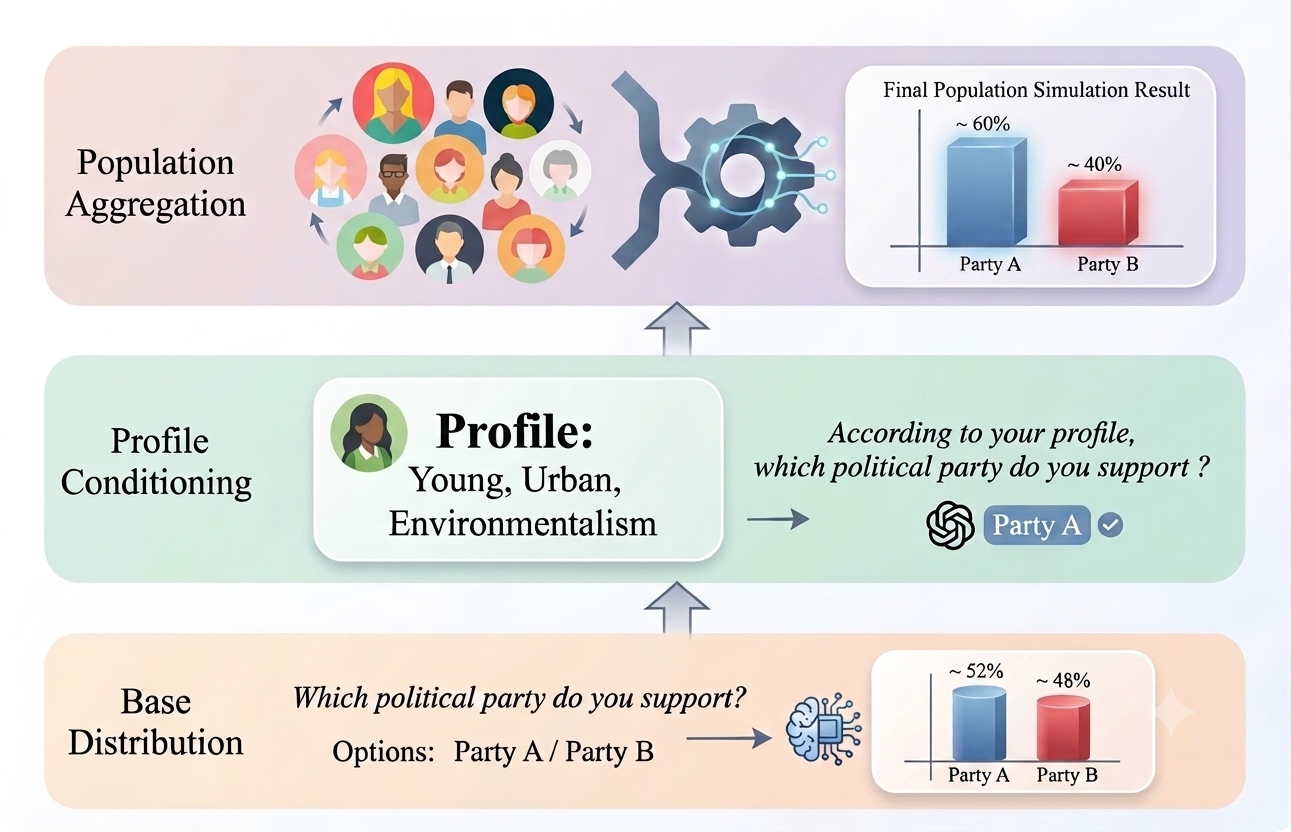} 
    \caption{LLM-based Human Behavior Simulation As a Hierarchical Structure.}
    \vspace{-0.5cm}
    \label{fig:intro}
\end{figure}

\begin{figure*}[htbp]
    \centering
    \includegraphics[width=0.88\textwidth]{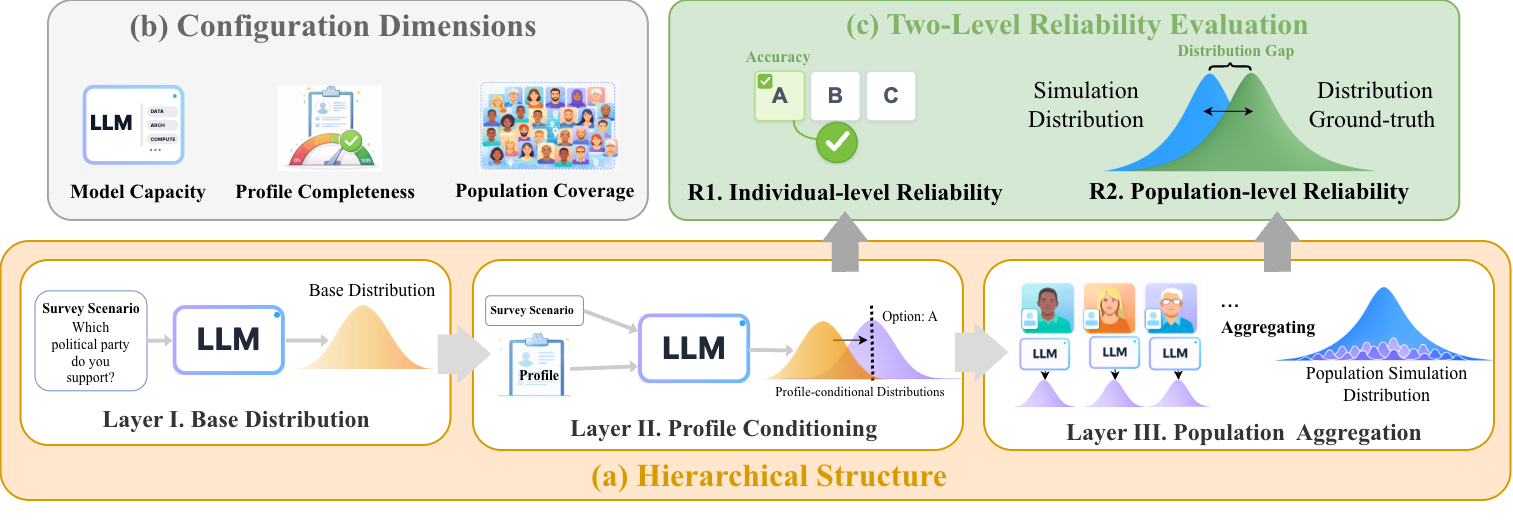}
    \caption{ReliMap Overview: Three-Layer Structure with Configurable Dimensions and Two-Level Reliability Evaluation.}
    \vspace{-10pt}
    \label{fig:main}
\end{figure*}
Recent years have seen growing efforts in this direction, spanning individual-level response prediction~\cite{aher2023usinglargelanguagemodels, hu2025simbenchbenchmarkingabilitylarge, wang2025sociobenchmodelinghumanbehavior} and population-level distribution matching~\cite{huang2025distributionshiftalignmenthelps,wang2025can}. Yet despite this breadth, the field remains fragmented. On the methodology side, some work focuses on prompting strategies and persona engineering to improve per-respondent simulation accuracy~\cite{bisbee2024synthetic, cho2024llm, moon2024virtualpersonaslanguagemodels}, while others fine-tune or align models to better reproduce empirical responses~\cite{chu2023languagemodelstrainedmedia, suh2025languagemodelfinetuningscaled, kolluri2025finetuningllmshumanbehavior}. On the evaluation side, some work assesses whether simulated individual responses match ground-truth labels~\cite{hewitt2024predicting}, while others examine whether simulated population distributions align with human survey distributions~\cite{huang2025distributionshiftalignmenthelps,wang2025can}. These lines of work each operate within a narrow scope, addressing one aspect of simulation in isolation. 
Consequently, it remains unclear how different configuration choices interact to drive reliability. Researchers lack the systematic understanding needed to diagnose and improve simulation quality.

We argue that this gap stems from treating simulation as a single-step process, when in fact it is hierarchical. Our key insight is that this hierarchical process unfolds in three layers, each building upon the previous one. As illustrated in Figure \ref{fig:intro}, these three layers are: (1) Base distribution: The model induces an intrinsic behavioral distribution shaped by pretraining and alignment, capturing its default tendency before any individual information is introduced. (2) Profile conditioning: This base distribution is then conditioned on individual profile information—such as demographics, values, or prior experiences—shifting behavioral tendencies toward those of specific personas. (3) Population aggregation: These individualized distributions are finally aggregated across a sampled population, where coverage and sampling strategy determine how faithfully the simulation reflects the target population. Building on this structure, we introduce \textbf{ReliMap} to make reliability analysis more structured and interpretable. Specifically, we systematically examine three key configuration dimensions—model capacity, profile completeness, and population coverage—and evaluate their joint effects at both the individual level (R1) and the population level (R2).

Guided by ReliMap, we conduct experiments on four simulation tasks with eleven LLMs. All models exhibit substantial population-level distributional bias without profile conditioning, with TVD varying widely across models and tasks. Our analysis reveals six findings: (1) Richer profiles improve R1 with diminishing returns. (2) Profile scaling behavior is strongly model-dependent: larger models benefit more from additional profile information. (3) Predictive information is highly concentrated in a small subset of attributes. (4) Attribute informativeness matters more than attribute count. (5) R1 gains do not reliably transfer to R2: individual accuracy improvements can coincide with worsening population-level alignment, revealing a fundamental decoupling between the two levels. (6) Population coverage reduces variance but not systematic bias, with R2 stabilizing at around 50–100 individuals. Together, these findings deepen our understanding of what drives reliability in LLM-based human behavior simulation.

In summary, our contributions are threefold: (1) We introduce ReliMap, which transforms reliability from a single score into a decomposable analysis across three layers, three configuration dimensions, and two evaluation levels, providing a principled framework for understanding and diagnosing LLM-based human behavior simulation. (2) We conduct systematic experiments across four simulation tasks and eleven LLMs, providing a comprehensive empirical examination of how configuration choices shape reliability at each simulation layer. (3) We reveal six findings that fundamentally advance our understanding of simulation reliability, exposing the limitations of current approaches and offering new insights into what it takes to achieve reliable LLM-based human behavior simulation.

\section{ReliMap}
\subsection{Hierarchical Structure}
\label{sec:pipeline}
As illustrated in Figure \ref{fig:main}(a), we characterize LLM-based human behavior simulation as a three-layer generative process, where each layer corresponds to a natural stage of how simulation is produced.

\paragraph{Layer I: Base distribution.} The first layer captures the model’s default responses without any personal profile information. It is the baseline distribution induced under a given scenario. This distribution is mainly shaped by the model architecture, pretraining data, and alignment strategy. Formally,
let $f_{\psi}$ denote a pretrained LLM with parameters $\psi$.
Given a scenario $s$, we define the base distribution as: 
\begin{equation}
\mathbb{P}_s
\;\triangleq\;
P_{f_{\psi}}(y \mid s).
\end{equation}

\begin{figure*}[t]
  \centering
  \includegraphics[width=\linewidth]{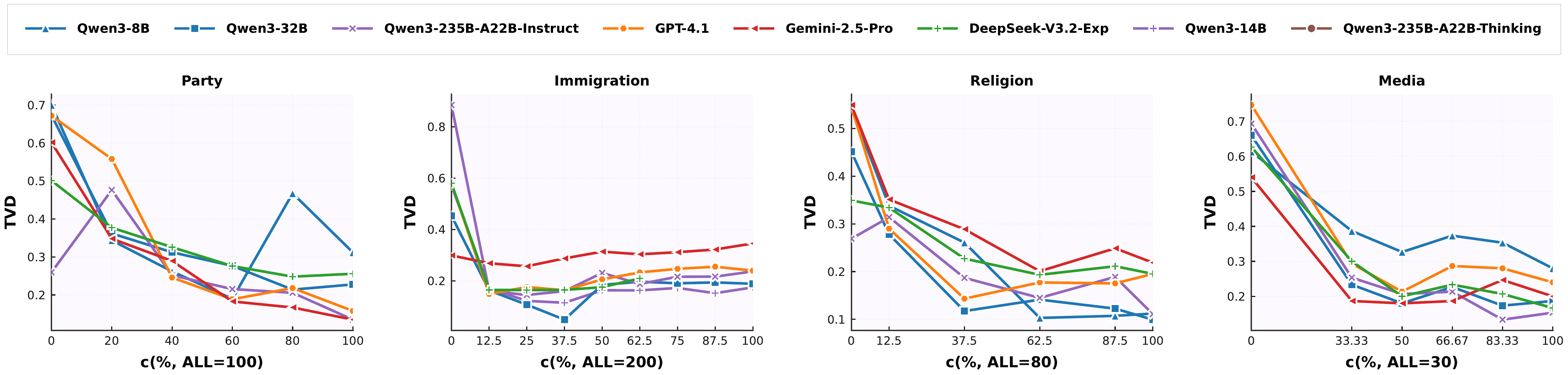}
  \caption{Effect of profile completeness on population simulation reliability.}
  \label{fig:tvd_emd_m}
\end{figure*}

\begin{table}[ht]
\centering
\footnotesize
\sisetup{
  table-number-alignment=right,
  table-format=1.3,
  detect-weight=true,
  detect-family=true
}
\setlength{\tabcolsep}{3pt}
\renewcommand{\arraystretch}{0.92}
\begin{tabular}{@{}l S S S S@{}}
\toprule
Model & {Party} & {Immig.} & {Relig.} & {Media} \\
\midrule
\multicolumn{5}{@{}l}{\textbf{Proprietary}} \\
GPT-4.1         & 0.670 & 0.368 & 0.542 & 0.747 \\
Gemini-2.5-Pro  & 0.589 & 0.297 & 0.549 & 0.527 \\
\midrule
\multicolumn{5}{@{}l}{\textbf{Open-source: DeepSeek}} \\
DeepSeek-V3.2-Exp & 0.483 &\bfseries{\num{0.128}} & 0.342 & 0.613 \\
DeepSeek-R1     & 0.664 & 0.579 & 0.345 & \textbf{0.480} \\
\midrule
\multicolumn{5}{@{}l}{\textbf{Open-source: Qwen3}} \\
Qwen3-8B                  & 0.689 & 0.304 & 0.540 & 0.620 \\
Qwen3-14B                 & 0.722 & 0.326 & 0.500 & 0.578 \\
Qwen3-32B                 & 0.664 & 0.408 & 0.474 & 0.653 \\
Qwen3-30B-A3B-Instruct    & 0.378 & 0.396 & 0.403 & 0.513 \\
Qwen3-30B-A3B-Thinking    & 0.527 & 0.181 & 0.443 & 0.647 \\
Qwen3-235B-A22B-Instruct  &\bfseries{\num{0.271}} & 0.482 & 0.263 & 0.684 \\
Qwen3-235B-A22B-Thinking  & 0.475 & 0.578 &\bfseries{\num{0.173}} & \textbf{0.831} \\
\bottomrule
\end{tabular}
\caption{TVD$\downarrow$ between LLM-induced base response distributions and human survey distributions (Layer I analysis; motivates the profile-conditioning study of Findings 1--5).}
\label{tab:stage1_tvd}
\end{table}

\paragraph{Layer II: Profile conditioning.} Building on the baseline distribution, the model is provided with an individual profile, such as geographic information, psychological tendencies, and past behaviors, yielding a profile-conditioned distribution. Formally,
given an individual profile $x$, we define the profile-conditional outcome distribution as 
\begin{equation}
\mathbb{P}_{s,x}
\;\triangleq\;
P_{f_{\psi}}(y \mid s, x).
\end{equation}

\begin{figure*}[!t]
  \centering

  \begin{subfigure}{\textwidth}
    \centering
    \includegraphics[width=\linewidth]{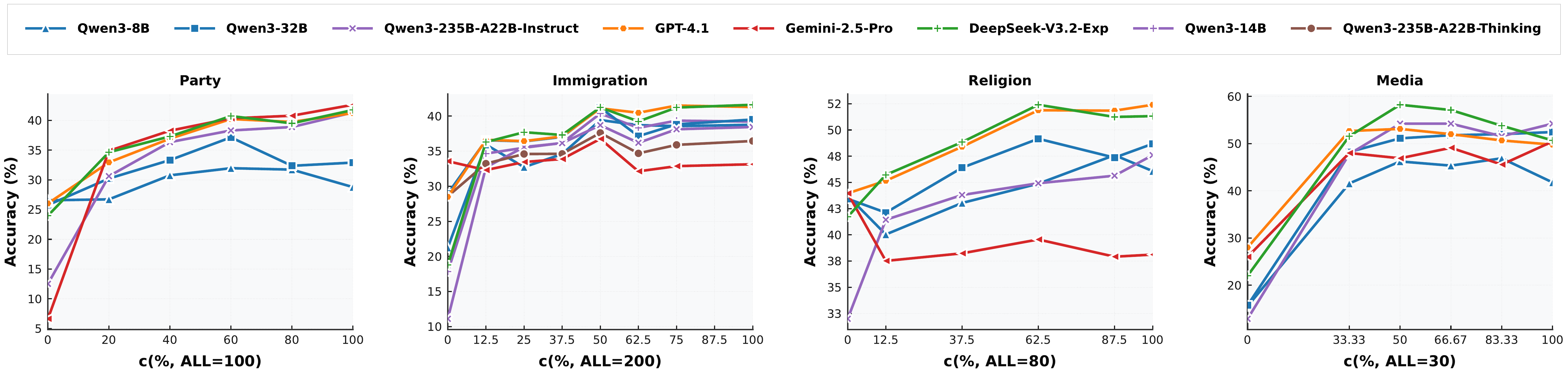}
    \vspace{-0.6cm}
    \caption{(a) $c$-Acc curves across model families}
    \label{fig:m_acc_curve_1}
  \end{subfigure}

  \vspace{0.1em}

  \begin{subfigure}{\textwidth}
    \centering
    \includegraphics[width=\linewidth]{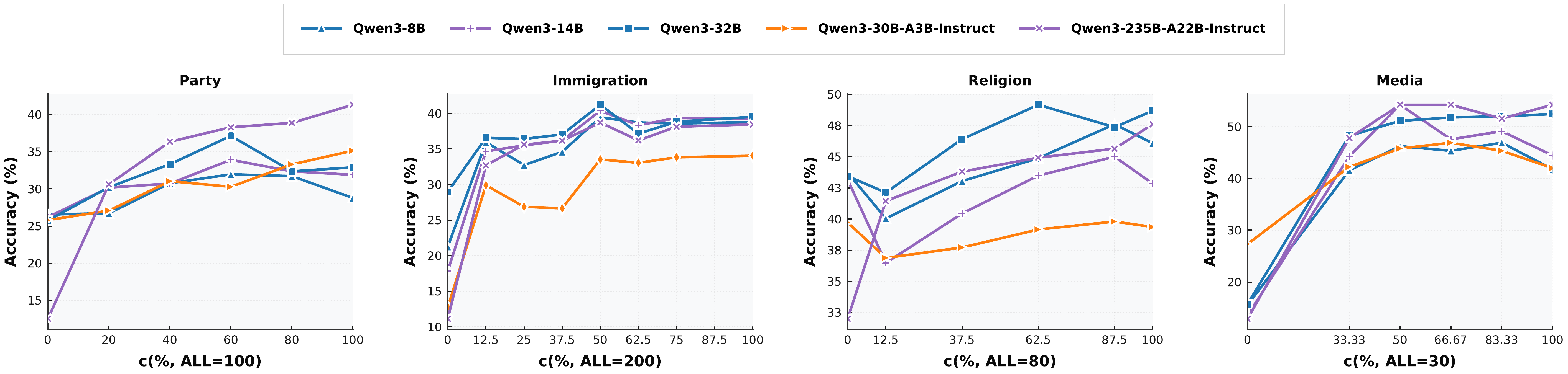}
    \vspace{-0.6cm}
    \caption{(b) $c$-ACC curves across model scales}
    \label{fig:m_acc_curve_2}
  \end{subfigure}

  \vspace{0.1em}

  \begin{subfigure}{\textwidth}
    \centering
    \includegraphics[width=\linewidth]{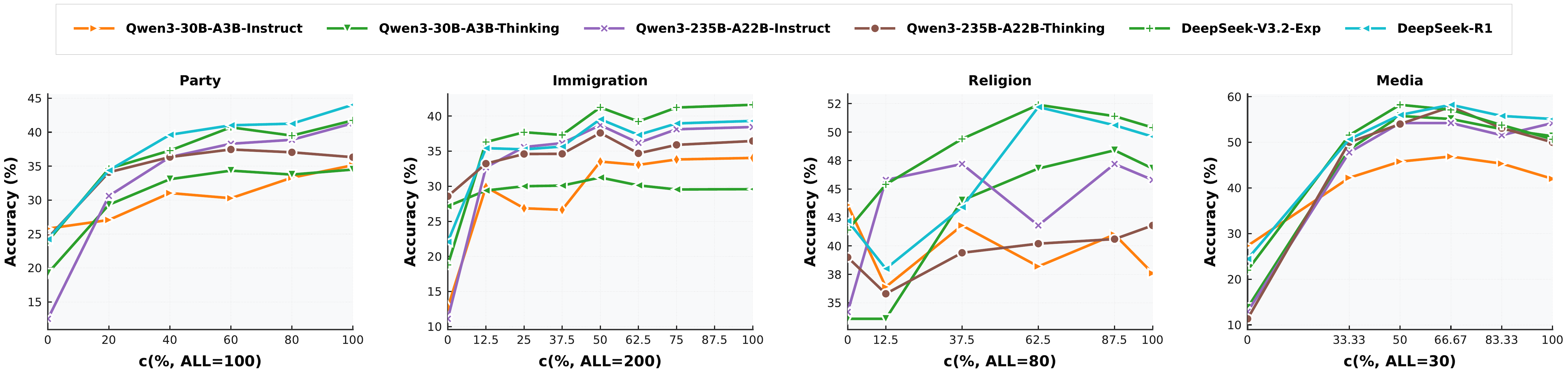}
    \vspace{-0.6cm}
    \caption{(c) $c$-ACC curves across reasoning variants}
    \label{fig:m_acc_curve_3}
  \end{subfigure}
    \vspace{-0.6cm}
  \caption{Changes in individual-level simulation accuracy as profile completeness $c$ increases across models. The x-axis $c$ denotes the profile attribute coverage rate, i.e., the percentage of attributes provided to the model out of all available attributes in the dataset. At each level of c, attributes are sampled randomly.}
  \label{fig:m_acc_curve}
\end{figure*}
\paragraph{Layer III: Population aggregation.}
To obtain population-level simulation results, the third layer aggregates the simulated outputs of sampled individuals to form a population distribution. Formally, we consider a finite population panel of N individuals
$\mathcal{X}_N = \{x_i\}_{i=1}^{N}$.
Aggregating individualized distributions of $\mathcal{X}_N$ yields the
population-aggregated outcome distribution:
\begin{equation}
\mathbb{P}_{s,\mathcal{X}_N}
\;\triangleq\;
\frac{1}{N} \sum_{i=1}^{N} P_{f_{\psi}}(y \mid s, x_i).
\label{eq:sim_marginal}
\end{equation}
Through the sequential generation and propagation from Layer I to Layer III, a standard LLM-based human behavior simulation can be completed.

\subsection{Configuration Dimensions}
We focus on three configuration dimensions that collectively capture the most critical controllable factors in the simulation setup, rather than implementation-level choices such as prompt templates or decoding strategies. As shown in Figure~\ref{fig:main}(b), three configuration dimensions are:

\paragraph{Model capacity ($\mathcal{M}$).} It concerns the choice of the base model, such as the model architecture and model size. This choice affects the simulation from Layer I and continues to propagate to later layers.
\paragraph{Profile completeness ($c$).} It captures how complete the available user information is. In principle, richer profiles should enable more accurate simulation of individual behavior. Yet in LLM-based simulation, the effect need not be linear. We therefore study how profile completeness shapes reliability, including when it matters and how strongly. It affects Layer II and propagates to Layer III.\footnote{We quantify $c$ as the proportion of attributes provided in a given profile relative to the total number of available attributes.}
\paragraph{Population coverage ($N$).}
 It refers to the coverage of the sample used in population simulation. Intuitively, broader coverage should lead to more reliable population-level results. However, this relationship needs to be quantified. We ask how much coverage is sufficient for a reliable population simulation, and how much additional gain we get from increasing coverage further. It affects Layer~III.

\subsection{Two-Level Reliability Evaluation}
As shown in Figure~\ref{fig:main}(c), we evaluate simulation reliability at two complementary levels: individual-level reliability (R1), which measures per-respondent prediction accuracy, and population-level reliability (R2), which measures how faithfully the aggregated simulation reproduces the true population distribution. These two levels capture distinct aspects of simulation quality and are not interchangeable. Intuitively, stronger R1 should translate into stronger R2, yet this coupling is not guaranteed. We therefore treat their joint evaluation as a necessary condition for a complete assessment of simulation reliability.

We state the R2 evaluation target precisely: R2 measures whether the model can reconstruct the response/preference distribution of a \emph{specified target group}. In our experiments, that target group is the retained analytical population after filtering respondents with complete and valid profile fields (see Appendix~\ref{appendix:data}), rather than the full original survey population, and we do not apply survey weights.

\paragraph{Theoretical Insight}
The three-layer structure, configuration dimensions, and two-level evaluation together form ReliMap. A key theoretical insight underlies this framework: population-level mismatch arises from multiple independent sources across simulation layers, and reducing one source does not automatically compensate for the others. Full derivations are provided in Appendix~\ref{appendix:theory}.

\section{Experiments}
\label{sec:experiments}
\subsection{Setup}
\paragraph{Data.}

We conduct experiments on four datasets. The first three are established survey datasets: partisan preference (Party) from the European Social Survey~\cite{ESS11int94:online}, immigration attitude (Immigration) from the World Values Survey~\cite{WVSDatab13:online}, and religious stance (Religion) from SocioBench~\cite{wang2025sociobenchmodelinghumanbehavior}. The fourth, Media, differs from the others in two ways. It is a Chinese-language dataset, and it captures a real-world dynamic event—public attitude toward the Luckin Coffee stock collapse—rather than a structured survey. We collect it from a public social media corpus.\footnote{Raw data sourced from \url{https://giser2000.github.io/geodata.github.io/\#download}.} More details are in Appendix~\ref{appendix:data}.

\paragraph{Metric.}
We evaluate simulation reliability at two levels. For R1, we use accuracy (ACC), which directly measures whether the model's predicted response matches the ground-truth individual behavior. For R2, we use total variation distance (TVD), which quantifies the gap between the simulated and ground-truth population distributions. We choose ACC and TVD because they directly match our discrete-choice R1/R2 setup and are widely adopted in prior work; we additionally verify that the main R2 trends remain consistent under KL divergence, Jensen--Shannon divergence (JSD), and Earth Mover's Distance (EMD), as reported in Appendix~\ref{app:alt_metrics}.

\paragraph{Model.}
We evaluate eleven LLMs covering a diverse range of model families, scales, and reasoning paradigms. Proprietary models include GPT-4.1~\cite{GPT-4.1} and Gemini-2.5-Pro~\cite{comanici2025gemini}. Open-source models include DeepSeek-V3.2-Exp~\cite{DeepSeek-v3.2} and DeepSeek-R1~\cite{guo2025deepseek}, as well as seven models from the Qwen3 family~\cite{yang2025qwen3} spanning different parameter scales and instruct/thinking variants. This diversity ensures that our findings are not specific to any single model family or architecture.

\subsection{Layer I — Base distribution}

In Layer I, we examine the model's base behavioral tendency without profile conditioning. We provide no profile information and only input the task context, making one prediction for each instance in the full dataset. Table~\ref{tab:stage1_tvd} shows the results.

TVD scores exhibit substantial variation across models and tasks. Even within a single task, models can differ markedly in distributional error—for example, on Party, the TVD of Qwen3-14B and Qwen3-235B-A22B-Instruct differs by approximately 0.45. Moreover, performance shows no consistent correlation with model scale or general capacity, suggesting that population-level distribution alignment constitutes a distinct capacity dimension. Crucially, all models exhibit non-trivial TVD across all tasks, indicating that profile conditioning is necessary to avoid systematic distributional bias. 


\subsection{Layer II — Profile Conditioning}
We investigate the impact of \textit{Model Capacity} and \textit{Profile Completeness}\footnote{Unless otherwise specified, attributes at each level of $c$ are randomly sampled by default.}  on R1, and obtained the following findings.
\paragraph{Finding 1: More Profile Context Helps, but Saturates Gradually}
We progressively increase the number of profile attributes and observe the corresponding changes in R1. The results are shown in Figure \ref{fig:m_acc_curve}. We find that across four tasks, ACC increases with the number of attributes, but with diminishing marginal gains. This aligns with our intuition that richer user profiles provide models with more personal characteristics, thereby facilitating behavioral simulation. The diminishing returns can be attributed to two factors. On the one hand, newly added attributes may be uninformative or redundant relative to existing ones. On the other hand, processing denser profiles requires stronger contextual understanding and reasoning from the model.

\paragraph{Finding 2: Profile Scaling Behavior is Strongly Model-Dependent}
To test whether the diminishing returns reflect a capacity bottleneck, we compare how different models scale with profile completeness.

\textit{Model Families.} As shown in Figure \ref{fig:m_acc_curve_1}, we find that \textbf{performance trends at higher $c$ are model-dependent}. As $c$ grows large, some models’ accuracy converges to a similar level and continues to increase such as Gemini-2.5-Pro, DeepSeek-V3.2-Exp, GPT-4.1 and Qwen3-235B-A22B-Instruct for \textit{Party}, while others plateau at a lower level such as Qwen3-8B and Qwen3-32B, creating a clear performance split. GPT-4.1 and DeepSeek-V3.2-Exp consistently perform best across all three tasks; as
$c$ increases, their ACC continues to improve substantially, indicating strong potential for further gains.

\textit{Model Size.} As shown in Figure~\ref{fig:m_acc_curve_2}, we observe a broadly consistent trend across the three tasks: \textbf{at the same profile completeness $c$, larger models tend to achieve higher accuracy and continue to improve as $c$ increases.} 
Within the Qwen3 family, Qwen3-32B consistently outperforms Qwen3-14B; under the MoE setting, Qwen3-235B-A22B-Instruct also generally outperforms Qwen3-30B-A3B-Instruct.

\textit{Instruct and Thinking variants.} Figure~\ref{fig:m_acc_curve_3} compares Instruct and Thinking models. We find no consistent evidence that Thinking models outperform Instruct models. Although Thinking models have been shown to demonstrate superior reasoning on mathematical and logical tasks, their reasoning abilities may differ from those needed for behavioral simulation. Simulating human behavior requires understanding social contexts, individual traits, and behavioral patterns that are potentially distinct from formal reasoning skills.

\paragraph{Finding 3: Behavioral Information Is Highly Concentrated Among Few Attributes}

\begin{table}[t]
\centering
\caption{Attribute informativeness statistics supporting Finding 3. Gini measures ASPG concentration across attributes (0 = uniform, 1 = maximally concentrated). Top 20\% info and Bot 50\% info denote the cumulative predictive information contributed by the top 20\% and bottom 50\% of attributes, respectively. Info@c=50\% reports cumulative information captured at half profile completeness under random (rand.) and informativeness-weighted (wtd.) sampling.}
\label{tab:aspg_stats}
\resizebox{\columnwidth}{!}{%
\begin{tabular}{@{}lrrr@{}}
\toprule
 & \textbf{Party} & \textbf{Immigration} & \textbf{Religion} \\
\midrule
\# attributes & 100 & 200 & 76 \\
ASPG median & 0.017 & 0.010 & 0.014 \\
Gini & 0.511 & 0.503 & 0.443 \\
Top 20\% info & 53.5\% & 53.3\% & 49.7\% \\
Bot 50\% info & 15.0\% & 16.9\% & 21.7\% \\
\midrule
Info@$c=50\%$ (rand.) & 45\% & 44\% & 47\% \\
Info@$c=50\%$ (wted.) & 86\% & 83\% & 79\% \\
\bottomrule
\end{tabular}%
}
\end{table}

To examine why performance saturates as profile completeness increases, we analyze the predictive information carried by individual attributes. We define \textit{Attribute-Specific Predictive Gain (ASPG)} as
\begin{equation}
    \text{ASPG}(a) = \frac{I(a; Y)}{H(Y)},
\end{equation}
where $I(a;Y)$ is the mutual information between attribute $a$ and target label $Y$, and $H(Y)$ is the label entropy. ASPG measures how much uncertainty about the target behavior can be reduced by observing a single attribute.\footnote{ASPG analysis excludes the Media task, which uses unstructured historical posts rather than predefined attributes.}

As shown in Table~\ref{tab:aspg_stats}, ASPG distributions are strongly right-skewed across all tasks. The top 20\% of attributes contribute roughly half of the total predictive information, while the bottom 50\% contribute less than 22\%. Under the same completeness budget ($c=50\%$), informativeness-weighted sampling captures 79--86\% of total information, compared to only 44--47\% under random sampling. These results show that predictive information is unevenly distributed across attributes, highlighting the importance of attribute selection.

\paragraph{Finding 4: Attribute Informativeness Matters More Than Attribute Count}

\begin{table}[t]
\centering
\scriptsize
\setlength{\tabcolsep}{8pt}
\renewcommand{\arraystretch}{1.05}
\begin{tabular}{@{}llrrr@{}}
\toprule
\textbf{Model} & & \textbf{Party} & \textbf{Immigration} & \textbf{Religion} \\
\midrule
\multirow{3}{*}{GPT4.1} & $r_Q$ & 0.9177 & 0.8368 & \textbf{0.9604} \\
 & $r_I$ & 0.9928 & 0.9662 & 0.9607 \\
 & $\Delta$ & \textbf{+0.0751} & \textbf{+0.1294} & \textbf{+0.0003} \\
\cmidrule(lr){1-5}
\multirow{3}{*}{Gemini-2.5-Pro} & $r_Q$ & 0.7835 & -0.0386 & -0.5336 \\
 & $r_I$ & 0.9624 & 0.0539 & -0.7305 \\
 & $\Delta$ & \textbf{+0.1789} & \textbf{+0.0925} & -0.1969 \\
\cmidrule(lr){1-5}
\multirow{3}{*}{DeepSeek-V3.2-Exp} & $r_Q$ & 0.8708 & 0.7233 & 0.8830 \\
 & $r_I$ & 0.9885 & 0.9251 & 0.9786 \\
 & $\Delta$ & \textbf{+0.1177} & \textbf{+0.2018} & \textbf{+0.0956} \\
\cmidrule(lr){1-5}
\multirow{3}{*}{Qwen3-235B-A22B} & $r_Q$ & 0.8562 & 0.6770 & 0.8485 \\
 & $r_I$ & 0.9898 & 0.9082 & 0.9654 \\
 & $\Delta$ & \textbf{+0.1336} & \textbf{+0.2312} & \textbf{+0.1169} \\
\cmidrule(lr){1-5}
\multirow{3}{*}{Qwen3-32B} & $r_Q$ & 0.6508 & 0.7117 & 0.8588 \\
 & $r_I$ & 0.8572 & 0.8911 & 0.8446 \\
 & $\Delta$ & \textbf{+0.2064} & \textbf{+0.1794} & -0.0142 \\
\cmidrule(lr){1-5}
\multirow{3}{*}{Qwen3-8B} & $r_Q$ & 0.6016 & 0.7355 & 0.8025 \\
 & $r_I$ & 0.7590 & 0.9132 & 0.6217 \\
 & $\Delta$ & \textbf{+0.1574} & \textbf{+0.1777} & -0.1808 \\
\midrule
\multirow{3}{*}{\textbf{Mean}} & $r_Q$ & 0.7801 & 0.6374 & 0.6366 \\
 & $r_I$ & 0.9250 & 0.8126 & 0.6067 \\
 & $\Delta$ & \textbf{+0.1449} & \textbf{+0.1752} & -0.0299 \\
\bottomrule
\end{tabular}
\caption{Pearson correlations between accuracy and two profile completeness metrics, supporting Finding 4. $r_Q$: quantity-based completeness; $r_I$: information-based completeness. $\Delta = r_I - r_Q$.}
\label{tab:pearson}
\end{table}

Table~\ref{tab:pearson} compares two ways of measuring the correlation between profile completeness and ACC: $r_Q$ measures the Pearson correlation between the proportion of provided attributes and ACC, and $r_I$ measures the Pearson correlation between the cumulative ASPG of provided attributes and ACC. Across most models and tasks, $r_I$ consistently exceeds $r_Q$. On average, information-based completeness improves the correlation with accuracy by \textbf{+0.1449} on \textit{Party} and \textbf{+0.1752} on \textit{Immigration}, indicating that the predictive information carried by attributes matters more than their count.

The trend is particularly strong for larger models such as GPT-4.1 and DeepSeek-V3.2-Exp, which achieve near-perfect correlations between $r_I$ and accuracy on several tasks. In contrast, the improvement is weaker or negative on \textit{Religion}, suggesting that the usefulness of informative attributes depends on task characteristics. Although ASPG relies on ground-truth labels and may not be directly applicable in real-world deployment, the results demonstrate that attribute quality is generally more important than attribute quantity.

We note that ASPG is a \emph{single-attribute} statistic: it measures how much predictive information one attribute carries about the target label in the data distribution, and being based on mutual information, it is strictly non-negative. ASPG therefore cannot by itself determine whether adding a particular attribute to an LLM prompt will help or hurt simulation performance. The performance degradation observed at some completeness levels (e.g., Table~\ref{tab:delta_acc_tvd}) is better understood as an effect of how the model jointly uses multiple attributes—individually informative attributes may still hurt performance when combined, due to redundancy, conflicting cues, longer-context interference, or model misinterpretation—rather than as evidence that any specific attribute is intrinsically harmful.

\subsection{Layer III — Population Aggregation.}
In Layer III, we evaluate whether improvements in individual-level simulation accuracy from Layer II reliably propagate to population-level distributional alignment. We analyze two questions: (i) how profile completeness gains in Layer II translate to R2(Finding 5), and (ii) how population coverage independently affects R2(Finding 6).

\paragraph{Finding 5: R1 Gains Do Not Reliably Transfer to R2}

We track R2 as Layer~II profile completeness $c$ increases in Figure~\ref{fig:tvd_emd_m}. R2 improves with $c$ but with diminishing returns: as profile completeness increases, the aggregated distribution moves closer to the empirical one, but the marginal gain shrinks as $c$ grows large and R1 begins to saturate. This overall trend, however, reflects only the average trajectory and does not guarantee that each incremental step in c produces consistent gains at both levels.

\begin{figure*}[t]
  \centering
    \includegraphics[width=\linewidth]{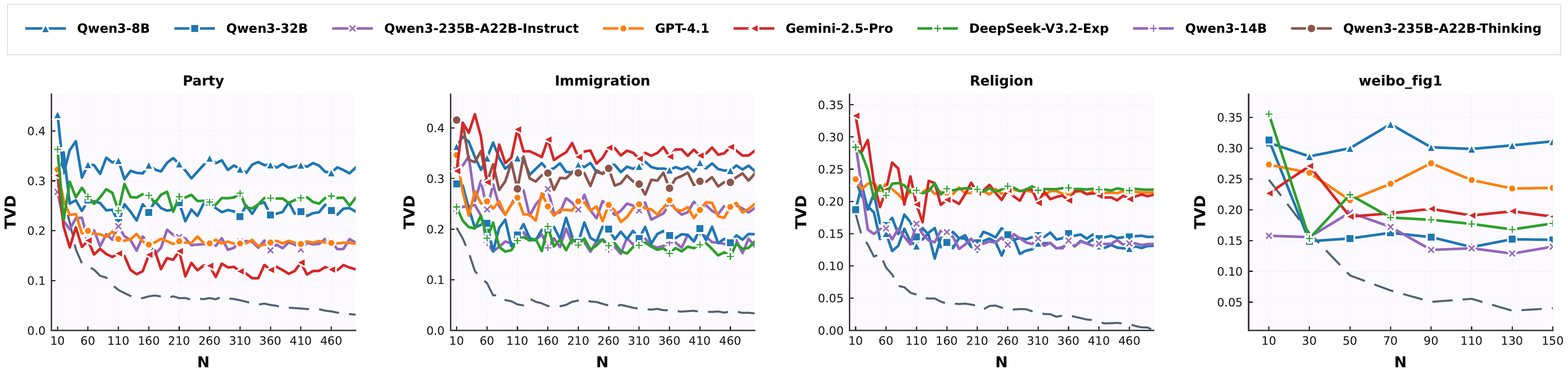}
  \caption{Population simulation reliability versus population coverage. TVD between the aggregated simulated distribution and the target population distribution as a function of the number of simulated individuals $N$. The dashed curve shows the GT sampling baseline: human ground-truth responses aggregated at varying sampling sizes, serving as a reference for the sampling variance of real human responses. The $c$ is fixed at 100\%.}
  \vspace{-10pt}
  \label{fig:tvd_n_curve}
\end{figure*}

\begin{table}[t]
\centering
\caption{Comparison of model performance across four datasets with varying profile completeness $c$. $\Delta$ACC and $\Delta$TVD denote the relative percentage change with respect to the previous profile-completeness level (i.e., the preceding row in each block). \textcolor{cbblue}{Blue} values mark steps where ACC increases while TVD also increases (R1 improves, R2 worsens); \textcolor{cborange}{Orange} values mark steps where both decrease (R1 worsens, R2 improves). In both cases R1 and R2 move in opposite directions, illustrating the decoupling in Finding 5.}
\label{tab:delta_acc_tvd}
\scriptsize
\setlength{\tabcolsep}{4pt}  
\begin{tabular}{@{}c|rrrrrr@{}}
\toprule
\multirow{2}{*}{\textbf{Dataset}} & \multicolumn{2}{c}{\textbf{GPT-4.1}} & \multicolumn{2}{c}{\textbf{DeepSeek-V3.2-Exp}} & \multicolumn{2}{c}{\textbf{Qwen3-235B-A22B}} \\
\cmidrule(lr){2-3} \cmidrule(lr){4-5} \cmidrule(lr){6-7}
 & $\Delta$ACC & $\Delta$TVD & $\Delta$ACC & $\Delta$TVD & $\Delta$ACC & $\Delta$TVD \\
\midrule
\multicolumn{7}{@{}l@{}}{\textbf{Party}} \\
20 & +26.50 & -55.52 & +44.45 & -43.66 & \textcolor{cbblue}{+144.57} & \textcolor{cbblue}{+5.91} \\
40 & +12.08 & -13.78 & +7.68 & -16.99 & +18.68 & -23.62 \\
60 & +8.94 & -31.88 & \textcolor{cbblue}{+9.17} & \textcolor{cbblue}{+6.33} & +5.39 & -39.73 \\
80 & -1.19 & +14.64 & \textcolor{cborange}{-3.00} & \textcolor{cborange}{-1.33} & \textcolor{cbblue}{+1.51} & \textcolor{cbblue}{+42.54} \\
100 & +3.72 & -22.73 & \textcolor{cbblue}{+5.67} & \textcolor{cbblue}{+5.12} & +6.28 & -18.26 \\
\midrule
\multicolumn{7}{@{}l@{}}{\textbf{Immig.}} \\
12.5 & +28.51 & -56.24 & -3.76 & +15.78 & +194.15 & -77.47 \\
25 & -0.52 & +3.27 & \textcolor{cbblue}{+3.56} & \textcolor{cbblue}{+2.90} & \textcolor{cbblue}{+8.84} & \textcolor{cbblue}{+10.58} \\
37.5 & +1.87 & -13.44 & +1.26 & -1.08 & +1.57 & -4.74 \\
50 & +10.76 & -19.93 & +8.48 & -40.87 & +7.06 & -15.16 \\
62.5 & -1.53 & +24.17 & -12.58 & +65.56 & -6.49 & +42.34 \\
75 & +2.55 & -3.15 & +2.30 & -3.15 & +5.34 & -5.56 \\
87.5 & -1.74 & +13.42 & \textcolor{cbblue}{+2.28} & \textcolor{cbblue}{+3.07} & +1.31 & -5.46 \\
100 & +1.25 & -3.75 & -1.43 & +3.15 & -0.47 & +7.48 \\
\midrule
\multicolumn{7}{@{}l@{}}{\textbf{Relig.}} \\
12.5 & +2.73 & -56.51 & +9.59 & -29.64 & +29.50 & -6.96 \\
37.5 & +7.17 & -25.37 & +6.82 & -15.12 & +5.69 & -34.86 \\
62.5 & +7.19 & -3.18 & \textcolor{cbblue}{+7.29} & \textcolor{cbblue}{+1.96} & \textcolor{cbblue}{+2.56} & \textcolor{cbblue}{+19.91} \\
87.5 & -0.08 & +9.62 & \textcolor{cborange}{-2.21} & \textcolor{cborange}{-0.76} & +1.60 & -16.73 \\
100 & \textcolor{cbblue}{+1.08} & \textcolor{cbblue}{+4.71} & +0.16 & -2.52 & +4.29 & -10.54 \\
\midrule
\multicolumn{7}{@{}l@{}}{\textbf{Med.}} \\
33.3 & +88.10 & -67.85 & +134.34 & -58.70 & +270.69 & -65.91 \\
50 & \textcolor{cbblue}{+0.84} & \textcolor{cbblue}{+2.77} & +12.93 & -21.92 & +13.49 & -10.48 \\
66.7 & \textcolor{cborange}{-2.09} & \textcolor{cborange}{-4.50} & -1.91 & +3.36 & +0.00 & -17.02 \\
83.3 & \textcolor{cborange}{-2.56} & \textcolor{cborange}{-7.54} & \textcolor{cborange}{-5.84} & \textcolor{cborange}{-15.22} & \textcolor{cborange}{-4.92} & \textcolor{cborange}{-20.51} \\
100 & -1.75 & +8.16 & -5.79 & +2.56 & \textcolor{cbblue}{+5.17} & \textcolor{cbblue}{+1.60} \\
\bottomrule
\end{tabular}
\end{table}

This overall improvement trend does not mean R1 gains reliably transfer to R2. As shown in Table~\ref{tab:delta_acc_tvd}, $\Delta\mathrm{ACC}$ and $\Delta\mathrm{TVD}$ can move in opposite directions across tasks, models, and completeness levels. We note that 26\% of incremental steps in Table~\ref{tab:delta_acc_tvd} show R1 and R2 moving in opposite directions.  

To illustrate how this decoupling manifests in practice, we examine two representative cases in Table~\ref{tab:case_study}. In Case 1 (Immigration, $c=12.5\% \to 25\%$), ACC improves but TVD worsens: as more profile information is provided, the model increasingly concentrates its predictions on \textit{Quite good} (bias: $+4.00\% \to +13.06\%$) while further under-predicting \textit{Neither good nor bad} and \textit{Very good}, pushing the predicted distribution away from the true population distribution. In Case 2 (Media, $c=66.7\% \to 83.3\%$), the reverse holds: the model's over-prediction of \textit{Bystander Entertainer} and under-prediction of \textit{Loyal Advocate} both partially correct, bringing the predicted distribution closer to the truth, but individual accuracy drops. Together, the cases reveal that TVD is driven by distributional shifts in a small subset of options, which can move independently of overall ACC.

\paragraph{Finding 6: Population Coverage Reduces Variance but Not Bias}
We study how population coverage affects R2 in Layer~III by varying the number of simulated individuals $N$ and measuring TVD between the aggregated simulated distribution and the target population distribution; see Figure~\ref{fig:tvd_n_curve}. As a reference, we include a ground-truth (GT) sampling baseline, obtained by aggregating actual human responses subsampled at varying population sizes $N$; this baseline characterizes the sampling variance of real human responses and isolates the error that remains even with perfect individual-level simulation. This distinction between variance and bias is further corroborated in Appendix \ref{app: more_exp}, where we jointly vary N and c.

\begin{table}[t]
\centering
\footnotesize
\setlength{\tabcolsep}{2.5pt}
\caption{Option-level prediction distributions for two decoupling cases (Qwen3-235B-A22B), illustrating Finding 5. GT\% denotes the occurrence rate of each option in the ground-truth human data. Bias values are colored by sign: \textcolor{darkred}{red} = over-prediction, \textcolor{darkblue}{blue} = under-prediction.
\textbf{Case 1} (Immigration): $c{=}12.5$\% ${\to} 25$\%, ACC$\uparrow$, TVD$\uparrow$. 
\textbf{Case 2} (Media): $c{=}66.7$\% ${\to} 83.3$\%, ACC$\downarrow$, TVD$\downarrow$.}
\label{tab:case_study}
\begin{tabular}{l|r|rr|rr}
\toprule
Option & GT\% & Pred\% & Bias & Pred\% & Bias \\
\midrule
\multicolumn{6}{c}{\textit{Case 1: Immigration ($c{=}12.5$\% ${\to} 25$\%)}} \\
\midrule
Very bad        & 7.49  & 5.81  & \textcolor{darkblue}{-1.68}  & 5.48  & \textcolor{darkblue}{-2.01}  \\
Quite bad       & 14.80 & 30.47 & \textcolor{darkred}{+15.67} & 23.49 & \textcolor{darkred}{+8.70}  \\
Neither         & 35.71 & 21.97 & \textcolor{darkblue}{-13.74} & 23.32 & \textcolor{darkblue}{-12.39} \\
Quite good      & 31.17 & 35.17 & \textcolor{darkred}{+4.00}  & 44.23 & \textcolor{darkred}{+13.06} \\
Very good       & 10.83 & 6.57  & \textcolor{darkblue}{-4.25}  & 3.47  & \textcolor{darkblue}{-7.35}  \\
\midrule
\multicolumn{6}{c}{\textit{Case 2: Media ($c{=}66.7$\% ${\to} 83.3$\%)}} \\
\midrule
Loyal Advocate        & 30.00 & 16.00 & \textcolor{darkblue}{-14.00} & 18.44 & \textcolor{darkblue}{-11.56} \\
Opportunist           & 26.67 & 24.89 & \textcolor{darkblue}{-1.78}  & 26.89 & \textcolor{darkred}{+0.22}  \\
Moral Outrage         & 12.00 & 12.22 & \textcolor{darkred}{+0.22}  & 11.33 & \textcolor{darkblue}{-0.67}  \\
Rational Evaluator    & 6.00  & 8.89  & \textcolor{darkred}{+2.89}  & 9.11  & \textcolor{darkred}{+3.11}  \\
Bystander Entertainer & 25.33 & 38.00 & \textcolor{darkred}{+12.67} & 34.22 & \textcolor{darkred}{+8.89}  \\
\bottomrule
\end{tabular}
\end{table}

Results show that population reliability improves quickly as coverage increases and then levels off. Model rankings are also unstable at small $N$, but become much more consistent once $N$ is large enough for TVD to stabilize. 
When $N$ is large, the remaining error is driven primarily by systematic distributional mismatch rather than sampling noise. 
As $N$ increases, the GT sampling baseline steadily decreases and approaches zero, whereas LLM-based simulations converge to a non-zero plateau. The persistent gap between the LLM curves and the GT baseline indicates that increasing $N$ reduces sampling variance but cannot eliminate the residual distributional mismatch.

\section{Related Work}
\begin{table*}[t]
\centering
\footnotesize
\setlength{\tabcolsep}{3pt}
\renewcommand{\arraystretch}{1.57}
\begin{tabular}{@{}lcccccc@{\hspace{22pt}}cccccccc@{}}
\toprule
 & \multicolumn{6}{c}{\textbf{Findings}} & \multicolumn{3}{c}{\textbf{Simulation Layers}} & \multicolumn{3}{c}{\textbf{Configurations}} & \multicolumn{2}{c}{\textbf{Evaluation}} \\
\cmidrule(lr){2-7} \cmidrule(lr){8-10} \cmidrule(lr){11-13} \cmidrule(lr){14-15}
\textbf{Work} & \textbf{F1} & \textbf{F2} & \textbf{F3} & \textbf{F4} & \textbf{F5} & \textbf{F6} & I & II & III & Cap. & Com. & Cov. & R1 & R2 \\
\midrule
\citet{hu2025simbenchbenchmarkingabilitylarge} & -- & R & -- & -- & -- & -- & $\times$ & \checkmark & \checkmark & \checkmark & $\times$ & $\times$ & $\times$ & \checkmark \\
\citet{wang2025sociobenchmodelinghumanbehavior} & -- & R & -- & -- & -- & -- & $\times$ & \checkmark & $\times$ & \checkmark & $\times$ & $\times$ & \checkmark & $\times$ \\
\citet{dominguezolmedo2024questioningsurveyresponseslarge} & -- & R & -- & -- & -- & -- & \checkmark & $\times$ & \checkmark & \checkmark & $\times$ & $\times$ & $\times$ & \checkmark \\
\citet{meister2024benchmarkingdistributionalalignmentlarge} & -- & -- & -- & -- & -- & -- & \checkmark & \checkmark & \checkmark & \checkmark & $\times$ & $\times$ & $\times$ & \checkmark \\
\citet{geng2024chameleons} & R & R & -- & -- & -- & -- & $\times$ & \checkmark & \checkmark & \checkmark & \checkmark & $\times$ & $\times$ & \checkmark \\
\citet{santurkar2023opinionslanguagemodelsreflect} & -- & -- & -- & -- & -- & -- & \checkmark & \checkmark & \checkmark & \checkmark & $\times$ & $\times$ & $\times$ & \checkmark \\
\citet{hwang2023aligning} & R & -- & -- & -- & -- & -- & \checkmark & \checkmark & $\times$ & \checkmark & \checkmark & $\times$ & \checkmark & $\times$ \\
\citet{qu2024performance} & R & -- & -- & -- & R & -- & $\times$ & \checkmark & \checkmark & \checkmark & \checkmark & $\times$ & \checkmark & \checkmark \\
\midrule
\textbf{ReliMap (ours)} & \checkmark & \checkmark & \checkmark & \checkmark & \checkmark & \checkmark & \checkmark & \checkmark & \checkmark & \checkmark & \checkmark & \checkmark & \checkmark & \checkmark \\
\bottomrule
\end{tabular}
\caption{Comparison with existing evaluation studies. (a) Coverage of our six findings F1--F6 (\S\ref{sec:experiments}): ``R'' = related prior evidence, ``--'' = not discussed. (b) Coverage of ReliMap's design space across simulation layers, configuration dimensions (Cap.: model capacity, Com.: profile completeness, Cov.: population coverage), and evaluation levels (R1/R2).}
\label{tab:relimap_comparison}
\end{table*}
\subsection{LLM-Based Simulation Evaluation} LLM-based human behavior simulation is increasingly used to generate survey responses, motivating benchmarks that target simulation reliability itself. \citet{santurkar2023opinionslanguagemodelsreflect} and \citet{durmus2024measuringrepresentationsubjectiveglobal} use real polls to test whether models match human viewpoint distributions across groups and countries. \citet{hu2025simbenchbenchmarkingabilitylarge} further treats simulation as a standalone capacity, covering diverse tasks with a unified, comparable protocol. \citet{meister2024benchmarkingdistributionalalignmentlarge} evaluates 43 LLMs against the American Community Survey and reports label bias and near-random responding, suggesting that the alignment scores may reflect closeness to uniform. However, existing evaluations still emphasize end-to-end scores, offering limited insight into where biases arise and how they propagate.
\subsection{Individual-level Simulation}
In individual-level simulation evaluation, prior work has explored methods for improving per-respondent simulation accuracy. \citet{bisbee2024synthetic} and \citet{tzachristas2025guided} show that zero-shot prompting is easy to deploy but can be highly sensitive to prompt design and sampling randomness, leading to unstable per-respondent accuracy. \citet{wang2025large} further finds that prompting-based simulation often underestimates the variance of human opinions, producing overly uniform outputs that obscure individual heterogeneity. Beyond prompting, \citet{wang2024large} studies conjoint analysis and proposes a statistical data-augmentation framework that integrates LLM-generated data with a small amount of human data to debias synthetic responses. In parallel, \citet{suh2025languagemodelfinetuningscaled} and \citet{cao2025specializinglargelanguagemodels} directly fine-tune LLMs on survey data to align token-level probabilities with the empirical distribution. However, these efforts largely conflate reliability with end-to-end improvement and offer limited insight into why simulations fail; ReliMap instead decomposes simulation into layers to localize error sources and guide targeted fixes.

\subsection{Population-level Distribution Alignment}
Beyond individual-level accuracy, recent work on pluralistic and distributional alignment treats group-level preference distributions as explicit training targets to better reflect heterogeneous values across users and groups. \citet{chakraborty2024maxminrlhfalignmentdiversehuman} targets group-level preference distributions for pluralistic alignment. \citet{melnyk2024distributionalpreferencealignmentllms} enforces distributional constraints, \citet{poddar2024personalizingreinforcementlearninghuman} models user/subgroup modes, and \citet{yao2025preferenceleftbehindgroup} adds group objectives to prevent minority collapse. ReliMap instead diagnoses \emph{survey-response} distribution shifts induced and propagated by configuration choices, not distribution matching as training.

\subsection{Comparison with Existing Evaluations}
\label{sec:comparison}
Table~\ref{tab:relimap_comparison} compares ReliMap with eight closely related evaluation studies. Some observations have prior support: profile/persona conditioning substantially affects simulation quality~\cite{geng2024chameleons,hwang2023aligning,qu2024performance}, and profile scaling varies across models~\cite{hu2025simbenchbenchmarkingabilitylarge,wang2025sociobenchmodelinghumanbehavior,dominguezolmedo2024questioningsurveyresponseslarge}. In contrast, F3 (concentration of predictive information), F4 (informativeness over attribute count), and F6 (variance--bias decomposition) are not systematically studied before; F5 is partially anticipated by \citet{qu2024performance}, which we turn into a layer-wise analysis. No existing study covers all three layers, three configuration dimensions, and both evaluation levels jointly; ReliMap unifies them to localize where reliability failures arise and propagate.

\section{Conclusion}
We introduce ReliMap, modeling LLM-based human behavior simulation as a three-layer generative process and evaluating reliability at individual and population levels across model capacity, profile completeness, and population coverage. Experiments on four tasks with eleven LLMs map reliability variation across layers and configurations. Individual-level gains do not transfer reliably to the population level, and larger coverage reduces variance but not systematic bias; reliable simulation requires coordinated improvement across layers.

\section*{Limitations}
\begin{itemize}
    [leftmargin=1em]
\item Our study focuses on human behavior simulation in survey-style tasks; our conclusions may not directly generalize to more complex settings such as interactive dialogue or multi-agent dynamics.

\item We evaluate individual-level reliability using accuracy, which directly reflects the discrete nature of survey responses but is a winner-takes-all metric that does not capture the subjectivity inherent in human decision-making. More principled alternatives remain an open direction for future work.

\item Our analysis is observational. While we identify how configuration choices affect reliability, the underlying mechanisms—such as why larger models benefit more from richer profiles—remain unexplored. Interpretability of LLM-based simulation is an important direction for future work.

\item Pretraining contamination cannot be fully ruled out, as the survey questions and public opinion data we use may appear in model pretraining corpora. This concern is most relevant to Layer I, where models generally perform poorly without profile conditioning, suggesting that contamination does not drive our main conclusions, which are comparative rather than absolute.

\end{itemize}

\bibliography{custom}

\begin{thebibliography}{38}
\providecommand{\natexlab}[1]{#1}

\bibitem[{Aher et~al.(2023)Aher, Arriaga, and Kalai}]{aher2023usinglargelanguagemodels}
Gati Aher, Rosa~I. Arriaga, and Adam~Tauman Kalai. 2023.
\newblock \href {https://arxiv.org/abs/2208.10264} {Using large language models to simulate multiple humans and replicate human subject studies}.
\newblock \emph{Preprint}, arXiv:2208.10264.

\bibitem[{Argyle et~al.(2023)Argyle, Busby, Fulda, Gubler, Rytting, and Wingate}]{Argyle_2023}
Lisa~P. Argyle, Ethan~C. Busby, Nancy Fulda, Joshua~R. Gubler, Christopher Rytting, and David Wingate. 2023.
\newblock \href {https://doi.org/10.1017/pan.2023.2} {Out of one, many: Using language models to simulate human samples}.
\newblock \emph{Political Analysis}, 31(3):337–351.

\bibitem[{Association(2022)}]{WVSDatab13:online}
The World Values~Survey Association. 2022.
\newblock \href {https://www.worldvaluessurvey.org/wvs.jsp} {Wvs database}.

\bibitem[{Binz et~al.(2025)Binz, Akata, Bethge, Br{\"a}ndle, Callaway, Coda-Forno, Dayan, Demircan, Eckstein, {\'E}ltet{\H{o}} et~al.}]{binz2025foundation}
Marcel Binz, Elif Akata, Matthias Bethge, Franziska Br{\"a}ndle, Fred Callaway, Julian Coda-Forno, Peter Dayan, Can Demircan, Maria~K Eckstein, No{\'e}mi {\'E}ltet{\H{o}}, and 1 others. 2025.
\newblock A foundation model to predict and capture human cognition.
\newblock \emph{Nature}, pages 1--8.

\bibitem[{Bisbee et~al.(2024)Bisbee, Clinton, Dorff, Kenkel, and Larson}]{bisbee2024synthetic}
James Bisbee, Joshua~D Clinton, Cassy Dorff, Brenton Kenkel, and Jennifer~M Larson. 2024.
\newblock Synthetic replacements for human survey data? the perils of large language models.
\newblock \emph{Political Analysis}, 32(4):401--416.

\bibitem[{Cao et~al.(2025)Cao, Liu, Arora, Augenstein, Röttger, and Hershcovich}]{cao2025specializinglargelanguagemodels}
Yong Cao, Haijiang Liu, Arnav Arora, Isabelle Augenstein, Paul Röttger, and Daniel Hershcovich. 2025.
\newblock \href {https://arxiv.org/abs/2502.07068} {Specializing large language models to simulate survey response distributions for global populations}.
\newblock \emph{Preprint}, arXiv:2502.07068.

\bibitem[{Chakraborty et~al.(2024)Chakraborty, Qiu, Yuan, Koppel, Huang, Manocha, Bedi, and Wang}]{chakraborty2024maxminrlhfalignmentdiversehuman}
Souradip Chakraborty, Jiahao Qiu, Hui Yuan, Alec Koppel, Furong Huang, Dinesh Manocha, Amrit~Singh Bedi, and Mengdi Wang. 2024.
\newblock \href {https://arxiv.org/abs/2402.08925} {Maxmin-rlhf: Alignment with diverse human preferences}.
\newblock \emph{Preprint}, arXiv:2402.08925.

\bibitem[{Cho et~al.(2024)Cho, Kim, and Kim}]{cho2024llm}
Suhyun Cho, Jaeyun Kim, and Jang~Hyun Kim. 2024.
\newblock Llm-based doppelg{\"a}nger models: leveraging synthetic data for human-like responses in survey simulations.
\newblock \emph{IEEE Access}.

\bibitem[{Chu et~al.(2023)Chu, Andreas, Ansolabehere, and Roy}]{chu2023languagemodelstrainedmedia}
Eric Chu, Jacob Andreas, Stephen Ansolabehere, and Deb Roy. 2023.
\newblock \href {https://arxiv.org/abs/2303.16779} {Language models trained on media diets can predict public opinion}.
\newblock \emph{Preprint}, arXiv:2303.16779.

\bibitem[{Comanici et~al.(2025)Comanici, Bieber, Schaekermann, Pasupat, Sachdeva, Dhillon, Blistein, Ram, Zhang, Rosen et~al.}]{comanici2025gemini}
Gheorghe Comanici, Eric Bieber, Mike Schaekermann, Ice Pasupat, Noveen Sachdeva, Inderjit Dhillon, Marcel Blistein, Ori Ram, Dan Zhang, Evan Rosen, and 1 others. 2025.
\newblock Gemini 2.5: Pushing the frontier with advanced reasoning, multimodality, long context, and next generation agentic capabilities.
\newblock \emph{arXiv preprint arXiv:2507.06261}.

\bibitem[{DeepSeek(2025)}]{DeepSeek-v3.2}
DeepSeek. 2025.
\newblock \href {https://api-docs.deepseek.com/news/news250929} {Deepseek-v3.2-exp release | deepseek api docs}.

\bibitem[{Dominguez-Olmedo et~al.(2024)Dominguez-Olmedo, Hardt, and Mendler-Dünner}]{dominguezolmedo2024questioningsurveyresponseslarge}
Ricardo Dominguez-Olmedo, Moritz Hardt, and Celestine Mendler-Dünner. 2024.
\newblock \href {https://arxiv.org/abs/2306.07951} {Questioning the survey responses of large language models}.
\newblock \emph{Preprint}, arXiv:2306.07951.

\bibitem[{Durmus et~al.(2024)Durmus, Nguyen, Liao, Schiefer, Askell, Bakhtin, Chen, Hatfield-Dodds, Hernandez, Joseph, Lovitt, McCandlish, Sikder, Tamkin, Thamkul, Kaplan, Clark, and Ganguli}]{durmus2024measuringrepresentationsubjectiveglobal}
Esin Durmus, Karina Nguyen, Thomas~I. Liao, Nicholas Schiefer, Amanda Askell, Anton Bakhtin, Carol Chen, Zac Hatfield-Dodds, Danny Hernandez, Nicholas Joseph, Liane Lovitt, Sam McCandlish, Orowa Sikder, Alex Tamkin, Janel Thamkul, Jared Kaplan, Jack Clark, and Deep Ganguli. 2024.
\newblock \href {https://arxiv.org/abs/2306.16388} {Towards measuring the representation of subjective global opinions in language models}.
\newblock \emph{Preprint}, arXiv:2306.16388.

\bibitem[{ESS~ERIC(2025)}]{ESS11int94:online}
Sikt ESS~ERIC. 2025.
\newblock \href {https://ess.sikt.no/en/datafile/242aaa39-3bbb-40f5-98bf-bfb1ce53d8ef} {Ess11 - integrated file, edition 4.0 | ess - sikt}.

\bibitem[{Geng et~al.(2024)Geng, He, and Trotta}]{geng2024chameleons}
Mingmeng Geng, Shiti He, and Roberto Trotta. 2024.
\newblock Are large language models chameleons?
\newblock In \emph{ICML 2024 Workshop on LLMs and Cognition}.

\bibitem[{Guo et~al.(2025)Guo, Yang, Zhang, Song, Zhang, Xu, Zhu, Ma, Wang, Bi et~al.}]{guo2025deepseek}
Daya Guo, Dejian Yang, Haowei Zhang, Junxiao Song, Ruoyu Zhang, Runxin Xu, Qihao Zhu, Shirong Ma, Peiyi Wang, Xiao Bi, and 1 others. 2025.
\newblock Deepseek-r1: Incentivizing reasoning capability in llms via reinforcement learning.
\newblock \emph{arXiv preprint arXiv:2501.12948}.

\bibitem[{Hewitt et~al.(2024)Hewitt, Ashokkumar, Ghezae, and Willer}]{hewitt2024predicting}
Luke Hewitt, Ashwini Ashokkumar, Isaias Ghezae, and Robb Willer. 2024.
\newblock Predicting results of social science experiments using large language models.
\newblock \emph{Preprint}.

\bibitem[{Hu et~al.(2025)Hu, Baumann, Lupo, Collier, Hovy, and Röttger}]{hu2025simbenchbenchmarkingabilitylarge}
Tiancheng Hu, Joachim Baumann, Lorenzo Lupo, Nigel Collier, Dirk Hovy, and Paul Röttger. 2025.
\newblock \href {https://arxiv.org/abs/2510.17516} {Simbench: Benchmarking the ability of large language models to simulate human behaviors}.
\newblock \emph{Preprint}, arXiv:2510.17516.

\bibitem[{Hu et~al.(2024)Hu, Kyrychenko, Rathje, Collier, van~der Linden, and Roozenbeek}]{hu2024generativelanguagemodelsexhibit}
Tiancheng Hu, Yara Kyrychenko, Steve Rathje, Nigel Collier, Sander van~der Linden, and Jon Roozenbeek. 2024.
\newblock \href {https://arxiv.org/abs/2310.15819} {Generative language models exhibit social identity biases}.
\newblock \emph{Preprint}, arXiv:2310.15819.

\bibitem[{Huang et~al.(2025)Huang, Li, and Shao}]{huang2025distributionshiftalignmenthelps}
Ji~Huang, Mengfei Li, and Shuai Shao. 2025.
\newblock \href {https://arxiv.org/abs/2510.21977} {Distribution shift alignment helps llms simulate survey response distributions}.
\newblock \emph{Preprint}, arXiv:2510.21977.

\bibitem[{Hwang et~al.(2023)Hwang, Majumder, and Tandon}]{hwang2023aligning}
EunJeong Hwang, Bodhisattwa~Prasad Majumder, and Niket Tandon. 2023.
\newblock Aligning language models to user opinions.
\newblock In \emph{Findings of the Association for Computational Linguistics: EMNLP 2023}, pages 5906--5919.

\bibitem[{Kolluri et~al.(2025)Kolluri, Wu, Park, and Bernstein}]{kolluri2025finetuningllmshumanbehavior}
Akaash Kolluri, Shengguang Wu, Joon~Sung Park, and Michael~S. Bernstein. 2025.
\newblock \href {https://arxiv.org/abs/2509.05830} {Finetuning llms for human behavior prediction in social science experiments}.
\newblock \emph{Preprint}, arXiv:2509.05830.

\bibitem[{Manning et~al.(2024)Manning, Zhu, and Horton}]{manning2024automatedsocialsciencelanguage}
Benjamin~S. Manning, Kehang Zhu, and John~J. Horton. 2024.
\newblock \href {https://arxiv.org/abs/2404.11794} {Automated social science: Language models as scientist and subjects}.
\newblock \emph{Preprint}, arXiv:2404.11794.

\bibitem[{Meister et~al.(2024)Meister, Guestrin, and Hashimoto}]{meister2024benchmarkingdistributionalalignmentlarge}
Nicole Meister, Carlos Guestrin, and Tatsunori Hashimoto. 2024.
\newblock \href {https://arxiv.org/abs/2411.05403} {Benchmarking distributional alignment of large language models}.
\newblock \emph{Preprint}, arXiv:2411.05403.

\bibitem[{Melnyk et~al.(2024)Melnyk, Mroueh, Belgodere, Rigotti, Nitsure, Yurochkin, Greenewald, Navratil, and Ross}]{melnyk2024distributionalpreferencealignmentllms}
Igor Melnyk, Youssef Mroueh, Brian Belgodere, Mattia Rigotti, Apoorva Nitsure, Mikhail Yurochkin, Kristjan Greenewald, Jiri Navratil, and Jerret Ross. 2024.
\newblock \href {https://arxiv.org/abs/2406.05882} {Distributional preference alignment of llms via optimal transport}.
\newblock \emph{Preprint}, arXiv:2406.05882.

\bibitem[{Moon et~al.(2024)Moon, Abdulhai, Kang, Suh, Soedarmadji, Behar, and Chan}]{moon2024virtualpersonaslanguagemodels}
Suhong Moon, Marwa Abdulhai, Minwoo Kang, Joseph Suh, Widyadewi Soedarmadji, Eran~Kohen Behar, and David~M. Chan. 2024.
\newblock \href {https://arxiv.org/abs/2407.06576} {Virtual personas for language models via an anthology of backstories}.
\newblock \emph{Preprint}, arXiv:2407.06576.

\bibitem[{OpenAI(2025)}]{GPT-4.1}
OpenAI. 2025.
\newblock \href {https://openai.com/index/gpt-4-1/} {Introducing gpt-4.1 in the api | openai}.

\bibitem[{Poddar et~al.(2024)Poddar, Wan, Ivison, Gupta, and Jaques}]{poddar2024personalizingreinforcementlearninghuman}
Sriyash Poddar, Yanming Wan, Hamish Ivison, Abhishek Gupta, and Natasha Jaques. 2024.
\newblock \href {https://arxiv.org/abs/2408.10075} {Personalizing reinforcement learning from human feedback with variational preference learning}.
\newblock \emph{Preprint}, arXiv:2408.10075.

\bibitem[{Qu and Wang(2024)}]{qu2024performance}
Yanchen Qu and Jian Wang. 2024.
\newblock Performance and biases of large language models in public opinion simulation.
\newblock \emph{Humanities and Social Sciences Communications}, 11(1):1--13.

\bibitem[{Santurkar et~al.(2023)Santurkar, Durmus, Ladhak, Lee, Liang, and Hashimoto}]{santurkar2023opinionslanguagemodelsreflect}
Shibani Santurkar, Esin Durmus, Faisal Ladhak, Cinoo Lee, Percy Liang, and Tatsunori Hashimoto. 2023.
\newblock \href {https://arxiv.org/abs/2303.17548} {Whose opinions do language models reflect?}
\newblock \emph{Preprint}, arXiv:2303.17548.

\bibitem[{Suh et~al.(2025)Suh, Jahanparast, Moon, Kang, and Chang}]{suh2025languagemodelfinetuningscaled}
Joseph Suh, Erfan Jahanparast, Suhong Moon, Minwoo Kang, and Serina Chang. 2025.
\newblock \href {https://arxiv.org/abs/2502.16761} {Language model fine-tuning on scaled survey data for predicting distributions of public opinions}.
\newblock \emph{Preprint}, arXiv:2502.16761.

\bibitem[{Tzachristas et~al.(2025)Tzachristas, Narayanan, and Antoniou}]{tzachristas2025guided}
Ioannis Tzachristas, Santhanakrishnan Narayanan, and Constantinos Antoniou. 2025.
\newblock Guided persona-based ai surveys: Can we replicate personal mobility preferences at scale using llms?
\newblock \emph{arXiv preprint arXiv:2501.13955}.

\bibitem[{Wang et~al.(2025{\natexlab{a}})Wang, Morgenstern, and Dickerson}]{wang2025large}
Angelina Wang, Jamie Morgenstern, and John~P Dickerson. 2025{\natexlab{a}}.
\newblock Large language models that replace human participants can harmfully misportray and flatten identity groups.
\newblock \emph{Nature Machine Intelligence}, pages 1--12.

\bibitem[{Wang et~al.(2025{\natexlab{b}})Wang, Pawlak, and Sivakumar}]{wang2025can}
Han Wang, Jacek Pawlak, and Aruna Sivakumar. 2025{\natexlab{b}}.
\newblock Can large language models simulate human responses? a case study of stated preference experiments in the context of heating-related choices.
\newblock \emph{arXiv preprint arXiv:2503.10652}.

\bibitem[{Wang et~al.(2025{\natexlab{c}})Wang, Zhao, Ni, and Wei}]{wang2025sociobenchmodelinghumanbehavior}
Jia Wang, Ziyu Zhao, Tingjuntao Ni, and Zhongyu Wei. 2025{\natexlab{c}}.
\newblock \href {https://arxiv.org/abs/2510.11131} {Sociobench: Modeling human behavior in sociological surveys with large language models}.
\newblock \emph{Preprint}, arXiv:2510.11131.

\bibitem[{Wang et~al.(2024)Wang, Zhang, and Zhang}]{wang2024large}
Mengxin Wang, Dennis~J Zhang, and Heng Zhang. 2024.
\newblock Large language models for market research: A data-augmentation approach.
\newblock \emph{arXiv preprint arXiv:2412.19363}.

\bibitem[{Yang et~al.(2025)Yang, Li, Yang, Zhang, Hui, Zheng, Yu, Gao, Huang, Lv et~al.}]{yang2025qwen3}
An~Yang, Anfeng Li, Baosong Yang, Beichen Zhang, Binyuan Hui, Bo~Zheng, Bowen Yu, Chang Gao, Chengen Huang, Chenxu Lv, and 1 others. 2025.
\newblock Qwen3 technical report.
\newblock \emph{arXiv preprint arXiv:2505.09388}.

\bibitem[{Yao et~al.(2025)Yao, Cai, Chuang, Yang, Jiang, Yang, and Hu}]{yao2025preferenceleftbehindgroup}
Binwei Yao, Zefan Cai, Yun-Shiuan Chuang, Shanglin Yang, Ming Jiang, Diyi Yang, and Junjie Hu. 2025.
\newblock \href {https://arxiv.org/abs/2412.20299} {No preference left behind: Group distributional preference optimization}.
\newblock \emph{Preprint}, arXiv:2412.20299.

\end{thebibliography}

\appendix

\section{Potential Risks}
This work examines LLM-based human behavior simulation in survey-style tasks and introduces ReliMap to diagnose reliability at both the individual and population levels. We use established survey datasets, collect no new personal data, and report results only in aggregate, which helps reduce privacy risks. Still, simulated responses can inherit biases from the underlying data and models, so they should not be interpreted as real public opinion. ReliMap is meant to support evaluation and auditing, rather than high-stakes use.

\section{Theoretical Insight}
\label{appendix:theory}
We provide theoretical insights into when and how the reliability of LLM-based human behavior simulation can converge to its ideal limit.
Rather than viewing reliability as a single score, we
show that it emerges from a multi-layer structure.

\paragraph{Multi-layer structure}
Let $s$ denote a task scenario, $u$ an individual, $x$ an individual profile, and $y$ a behavioral outcome.
Human behavior is assumed to arise from a structured process.
The target behavior distribution under scenario $s$ is
\begin{equation}
p^{\ast}(y \mid s)
=
\sum_{u} p^{\ast}(u)
\sum_{x} p^{\ast}(x \mid u)\,
p^{\ast}(y \mid s, x, u).
\end{equation}
An LLM-based simulator induces an approximate distribution of the same form,
\begin{equation}
p(y \mid s)
=
\sum_{u} p(u)
\sum_{x} p(x \mid u)\,
p(y \mid s, x, u),
\end{equation}
where deviations arise from imperfect modeling of individual sampling, profile construction,
and scenario-conditioned behavior generation.

\paragraph{Reliability gap.}
We define the population-level discrepancy under scenario $s$ as the $\ell_{1}$ distance
\begin{equation}
D(s)
\triangleq
\bigl\|p^{\ast}(\cdot \mid s)-p(\cdot \mid s)\bigr\|_{1}.
\end{equation}

\paragraph{Layer-wise error propagation.}
Consider a single $(u,x)$ term and define
\vspace{1mm}
\begin{equation}
\begin{aligned}
A^{\ast}&=p^{\ast}(y \mid s,x,u),\\[-1mm]
B^{\ast}&=p^{\ast}(x \mid u),\\[-1mm]
C^{\ast}&=p^{\ast}(u).
\end{aligned}
\end{equation}
and their simulator counterparts
\begin{equation}
A=p(y \mid s,x,u),\quad
B=p(x \mid u),\quad
C=p(u).
\end{equation}
The discrepancy contributed by this term is $\bigl|A^{\ast}B^{\ast}C^{\ast}-ABC\bigr|$.
By adding and subtracting intermediate terms and applying the triangle inequality, we have
\begin{equation}
\begin{aligned}
|A^{\ast}B^{\ast}C^{\ast}-ABC|
\le\;&
|B^{\ast}C^{\ast}|\,|A^{\ast}-A|
\\
&+
|AC^{\ast}|\,|B^{\ast}-B|
\\
&+
|AB|\,|C^{\ast}-C|.
\end{aligned}
\end{equation}
\vspace{0.07cm}

\paragraph{Bounding the overall error.}
For any fixed $y$, the difference between the target and simulated probabilities can be written as
\begin{equation}
p^{\ast}(y \mid s)-p(y \mid s)
=
\sum_{u}\sum_{x}\bigl(A^{\ast}B^{\ast}C^{\ast}-ABC\bigr).
\end{equation}
Applying the triangle inequality and summing over $y$ yields
\begin{equation}
\begin{aligned}
D(s)
&=
\sum_{y}\bigl|p^{\ast}(y \mid s)-p(y \mid s)\bigr|
\\
&\le
\sum_{y,u,x}\bigl|A^{\ast}B^{\ast}C^{\ast}-ABC\bigr|.
\end{aligned}
\end{equation}

Since all terms are probabilities in $[0,1]$, we can further upper bound each term by
\begin{equation}
|A^{\ast}B^{\ast}C^{\ast}-ABC|
\le
|A^{\ast}-A|+|B^{\ast}-B|+|C^{\ast}-C|.
\end{equation}

which gives the overall bound
\begin{equation}
D(s)
\le
\sum_{y}\sum_{u}\sum_{x}\Bigl(
\lvert A^{\ast}-A\rvert+\lvert B^{\ast}-B\rvert+\lvert C^{\ast}-C\rvert
\Bigr).
\end{equation}

\paragraph{Convergence conditions.}
The bound implies that $D(s)\to 0$ is guaranteed when the following component distributions converge:
\begin{equation}
\begin{aligned}
p(u) &\to p^{\ast}(u),\\[-1mm]
p(x \mid u) &\to p^{\ast}(x \mid u),\\[-1mm]
p(y \mid s,x,u) &\to p^{\ast}(y \mid s,x,u).
\end{aligned}
\end{equation}

This result shows that reliable simulation cannot be attained by optimizing any
single component in isolation.
Instead, reliability emerges from coordinated convergence across individual sampling,
profile construction, and scenario-conditioned behavior generation.

\section{Data Construction \& Statistics}
\label{appendix:data}
We conduct experiments on four tasks across two settings. The first setting comprises three established survey tasks: \textit{partisan preference}, \textit{immigration attitude}, and \textit{religious stance}. Task definitions are drawn from the European Social Survey (ESS; \textit{Party})~\cite{ESS11int94:online}, the World Values Survey (WVS; \textit{Immigration})~\cite{WVSDatab13:online}, and SocioBench (\textit{Religion})~\cite{wang2025sociobenchmodelinghumanbehavior}. For each dataset, we follow a three-step process. First, we select the survey questions that define the simulation tasks. Second, we preprocess the raw survey data by merging demographic attributes with historical survey responses to construct individual profiles. Finally, we filter the population to retain only respondents with complete and valid profile fields. Table~\ref{tab:task_settings_questions} reports statistics of the remaining valid data and the associated survey question for each task. Table~\ref{tab:task_options} lists the response options for each survey question.

The second setting introduces a real-world dynamic event: public attitude toward the Luckin Coffee stock collapse. It is in Chinese, further testing whether our findings generalize across languages and beyond traditional survey scenarios. Raw data were sourced from a publicly available social media dataset.\footnote{\url{https://giser2000.github.io/geodata.github.io/\#download}.} We filtered it for original posts containing the hashtag ``\#Luckin Coffee Collapse'' during the event period. Ground-truth labels were obtained by applying GPT-4o as an annotator to map users' original posts to the predefined attitude categories. Note that GPT-4o is used only for annotation, not for generating the target behavior: the target evidence consists of real user-authored posts about the event, so this construction is not circular. To verify label quality, we manually inspected 20 annotated examples and found 100\% annotation accuracy on this subset. Individual profiles were constructed from each user's 30 most recent historical posts prior to the event. Table~\ref{tab:task_settings_questions} includes statistics for this task alongside the three surveys.

\begin{table*}[!t]
\centering
\footnotesize  
\setlength{\tabcolsep}{4pt}
\renewcommand{\arraystretch}{1.25}
\begin{tabularx}{\textwidth}{@{}l >{\raggedright\arraybackslash}X >{\raggedright\arraybackslash}X >{\raggedright\arraybackslash}X >{\raggedright\arraybackslash}X@{}}
\toprule
\textbf{Item} & \textbf{Party} & \textbf{Immigration} & \textbf{Religious} & \textbf{Media} \\
\midrule
\texttt{profile\_attr\_num} & 100 & 200 & 80 & 30 \\
\texttt{population\_num} & 1,000 & 9,994 & 500 & 150 \\
\texttt{question} &
Which political party do you feel closest to? (Germany) &
How would you evaluate the impact of immigrants on your country's development? &
Do churches and religious organizations have too much or too little power? &
Analyze this user's attitude toward the `Luckin Coffee stock price crash' incident. \\
\bottomrule
\end{tabularx}
\caption{Task settings and question texts.}
\label{tab:task_settings_questions}
\end{table*}

\begin{table*}[!t]
\centering
\small
\setlength{\tabcolsep}{3pt}
\renewcommand{\arraystretch}{1.1}
\begin{tabular}{@{}>{\raggedright\arraybackslash}p{3.4cm} >{\raggedright\arraybackslash}p{3.4cm} >{\raggedright\arraybackslash}p{3.4cm} >{\raggedright\arraybackslash}p{3.4cm}@{}}
\toprule
\textbf{Party (options)} & \textbf{Immigration (options)} & \textbf{Religious (options)} & \textbf{Media (options)} \\
\midrule
1.\ CDU/CSU\newline
2.\ SPD\newline
3.\ The Left (Die Linke)\newline
4.\ Alliance 90/The Greens\newline
5.\ FDP\newline
6.\ AFD\newline
7.\ Free Voters\newline
8.\ dieBasis\newline
9.\ Die PARTEI
&
1.\ Very bad\newline
2.\ Quite bad\newline
3.\ Neither good nor bad\newline
4.\ Quite good\newline
5.\ Very good
&
1.\ Far too much power\newline
2.\ Too much power\newline
3.\ About the right amount\newline
4.\ Too little power\newline
5.\ Far too little power
&
1.\ Loyal Advocate\newline
2.\ Opportunist\newline
3.\ Moral Outrage\newline
4.\ Rational Evaluator\newline
5.\ Bystander Entertainer
\\
\bottomrule
\end{tabular}
\caption{Answer options for each task.}
\label{tab:task_options}
\end{table*}

\begin{figure}[!t]
\centering
\small
\begin{tcolorbox}[
    colback=white,
    colframe=black!60,
    boxrule=0.5pt,
    arc=2pt,
    left=4pt,
    right=4pt,
    top=3pt,
    bottom=3pt,
    boxsep=0pt,
    title={\textbf{Prompt: ISSP Survey Simulation}},
    fonttitle=\bfseries\small,
    coltitle=black,
    colbacktitle=gray!15,
    attach boxed title to top left={yshift=-2mm, xshift=4mm},
    boxed title style={arc=1pt, boxrule=0.3pt, colback=gray!15}
]
\setlength{\parskip}{0pt}
\setlength{\parsep}{0pt}
\setlength{\itemsep}{1pt}

{\footnotesize\textbf{Instruction:} You are participating in the International Social Survey Programme. Assume the role of a real individual with the following personal information. Fully immerse yourself in this persona and answer the question truthfully, based solely on the provided personal information and your previous answers to prior questions.}

\vspace{2pt}
{\footnotesize\textbf{Personal Information and Previous Answers:} \texttt{\{attributes\}}}

\vspace{2pt}
{\footnotesize\textbf{Question:} \texttt{\{question\}}}

\vspace{2pt}
{\footnotesize\textbf{Options:} \texttt{\{options\}}}

\vspace{2pt}
{\footnotesize\textbf{JSON Output:}}
\begin{lstlisting}[style=jsonstyle,basicstyle=\scriptsize\ttfamily,aboveskip=0pt,belowskip=0pt]
{"reason": "", "option": ""}
\end{lstlisting}

\vspace{2pt}
{\footnotesize\textbf{Requirements:}}
\begin{enumerate}[nosep,leftmargin=12pt,label=\arabic*.,topsep=0pt,itemsep=0.5pt]
    \item[\scriptsize 1.] {\scriptsize Answer based on personal information only; 6--10 sentence justification.}
    \item[\scriptsize 2.] {\scriptsize Choose best option; respond with number only.}
\end{enumerate}
\end{tcolorbox}
\caption{Prompt template for Party, Immigration, Religion.}
\label{fig:prompt_template}
\end{figure}

\begin{figure*}[!t]
\centering
\small
\begin{tcolorbox}[
    colback=white,
    colframe=black!60,
    boxrule=0.5pt,
    arc=2pt,
    left=4pt,
    right=4pt,
    top=3pt,
    bottom=3pt,
    boxsep=0pt,
    title={\textbf{Prompt: User Attitude Simulation}},
    fonttitle=\bfseries\small,
    coltitle=black,
    colbacktitle=gray!15,
    attach boxed title to top left={yshift=-2mm, xshift=4mm},
    boxed title style={arc=1pt, boxrule=0.3pt, colback=gray!15}
]
\setlength{\parskip}{0pt}
\setlength{\parsep}{0pt}
\setlength{\itemsep}{1pt}

{\footnotesize\textbf{Instruction:} You are a user attitude simulation model. Based on historical Weibo posts provided, simulate this user's likely attitude toward ``Luckin Coffee's stock price plummeted'' and determine which category best matches.}

\vspace{2pt}
{\footnotesize\textbf{Five Attitude Categories:}}
\begin{enumerate}[nosep,leftmargin=12pt,label=\arabic*.,topsep=0pt,itemsep=0.5pt]
    \item[\scriptsize 1.] {\scriptsize\textbf{Loyal Advocacy:} High emotional support; views Luckin as ``pride of domestic brands''; heartbreak/defensiveness on stock crash without blaming company.}
    \item[\scriptsize 2.] {\scriptsize\textbf{Opportunistic Exploitation:} No loyalty; only cares about prices/discounts/investment; neutral on fraud; motivation: ``getting a bargain'' or ``buying the dip.''}
    \item[\scriptsize 3.] {\scriptsize\textbf{Moral Indignation:} Strong moral outrage on corporate integrity; challenges ``national brand'' narrative; fraud is unforgivable.}
    \item[\scriptsize 4.] {\scriptsize\textbf{Belief Evaluation:} Distinguishes product from corporate behavior; may acknowledge quality but does not forgive fraud; rational, analytical stance.}
    \item[\scriptsize 5.] {\scriptsize\textbf{Entertainment Spectator:} No substantive stance; memes/jokes/viral quips (e.g., ``Sauce-flavored Latte becomes Limit-down Latte''); lighthearted, social interaction focused.}
\end{enumerate}

\vspace{2pt}
{\footnotesize\textbf{Input:} \texttt{\{posts\}} $\rightarrow$ \textbf{Event:} Luckin Coffee's stock price plummeted}

\vspace{2pt}
{\footnotesize\textbf{Output Rules:}}
\begin{enumerate}[nosep,leftmargin=12pt,label=\arabic*.,topsep=0pt,itemsep=0.5pt]
    \item[\scriptsize 1.] {\scriptsize Output exactly one valid JSON object only, no extra text/spaces/line breaks/Markdown.}
    \item[\scriptsize 2.] {\scriptsize JSON must contain: \texttt{analysis} (string) and \texttt{attitude\_id} (integer 1--5).}
    \item[\scriptsize 3.] {\scriptsize Do NOT use \texttt{```json} or code block markers. Output raw JSON directly.}
\end{enumerate}

\vspace{2pt}
{\footnotesize\textbf{JSON Output:}}
\begin{lstlisting}[style=jsonstyle,basicstyle=\scriptsize\ttfamily,aboveskip=0pt,belowskip=0pt]
{"analysis": "", "attitude_id": 0}
\end{lstlisting}
\end{tcolorbox}
\caption{Prompt template for Media Task (translated from Chinese).}
\label{fig:prompt_template_attitude}
\end{figure*}

\section{Experimental Details}

\subsection{Prompt Templates For Prediction}
Figure~\ref{fig:prompt_template} shows the prompt template used to elicit survey-style responses in the International Social Survey Programme setting.
Our design follows prior prompt-based survey simulation setups in \citet{wang2025sociobenchmodelinghumanbehavior}.
Each instance provides (i) an instruction to role-play as a real individual, (ii) the respondent’s personal attributes and any previous answers, (iii) the target question and its candidate options, and (iv) an explicit JSON output schema.
We require the model to justify its choice in 6--10 sentences based only on the provided attributes, and to return the selected option as a number in the \texttt{option} field.
This structured format standardizes generation across models and facilitates automatic parsing and evaluation. The prompt design for the Media task follows a similar structure, as shown in Figure \ref{fig:prompt_template_attitude}.

\subsection{Generation Settings}
We employ greedy decoding across all models. For Qwen3-8B, Qwen3-14B, and Qwen3-32B, we activate the \texttt{think} mode via \texttt{enable\_think=True}.
In profile-completeness experiments, we evaluate each completeness $c$ across \textbf{5 random seeds}, with each seed generating a distinct subset of profile attributes. Similarly, population-coverage experiments use \textbf{5 random seeds} for stochastic population sampling. All reported results represent averages over these replications.

\section{Illustrative Example of the Simulation Pipeline}
\label{app:example}
To clarify how the formal model is instantiated in practice, we provide a concrete example from the Party task. Table~\ref{tab:example} shows how the three key components—scenario $s$, individual $u$, and profile $x$—are populated in a real simulation instance.

In this framework, $u$ refers to a specific survey respondent (e.g., Respondent\_1). The scenario $s$ defines the task context, including the question and candidate options. The profile $x$ is a subset of attributes sampled from the full set of attributes available for $u$, which corresponds to $p(x \mid u)$ in our formulation. Studying how simulation performance varies as we include more or fewer attributes from $x$ allows us to quantify the effect of profile completeness on reliability.

Note that two users with identical attribute sets $x$ will indeed generate identical prompts. In practice, however, since we employ a rich set of high-dimensional sociodemographic features, cases where two distinct individuals share the same $u$ are extremely rare in our dataset and can be considered negligible.

\begin{table*}[h]
\centering
\small
\begin{tabular}{p{2cm}p{10cm}}
\toprule
\textbf{Component} & \textbf{Content} \\
\midrule
$s$ (scenario) & \textit{Question}: Which political party do you feel closest to? (Germany) \textit{Options}: 1. CDU/CSU, 2. SPD, 3. The Left (Die Linke), 4. Alliance 90/The Greens, 5. FDP, 6. AFD, 7. Free Voters, 8. dieBasis, 9. Die PARTEI \\
\midrule
$u$ (individual) & Respondent\_1 \\
\midrule
$x$ (profile) & \{``What the country needs most is loyalty towards its leaders.'': ``Disagree strongly'', ``How worried are you about climate change?'': ``Not at all worried'', ``Have you felt unfairly treated by the police because you are a man/woman?'': ``No'', ...\} \\
\midrule
Prompt & \texttt{\#\#\# Instruction:} You are participating in the International Social Survey Programme. Assume the role of a real individual with the following personal information. Fully immerse yourself in this persona and answer the question truthfully, based solely on the provided personal information and your previous answers to prior questions. \texttt{\#\#\# Personal Information and Previous Answers:} \{content from $x$\} \texttt{\#\#\# Question:} \{content from $s$\} \texttt{\#\#\# Options:} \{content from $s$\} \\
\midrule
Response & \{``reason'': ``Based on my previous answers, I am strongly in favor of gender equality in business leadership [...] SPD is also traditionally supportive of worker rights and moderate reforms, which matches my practical, balanced approach. Therefore, I feel closest to the SPD.'', ``option'': ``2''\} \\
\midrule
Ground Truth & 2 \\
\midrule
ACC & 1.0 \\
\bottomrule
\end{tabular}
\caption{A concrete example of the simulation pipeline for the Party task, illustrating how $s$, $u$, and $x$ are instantiated in practice.}
\label{tab:example}
\end{table*}

\section{More Experiments}
\label{app: more_exp}

\subsection{ASPG Distribution}
\begin{figure}[t]
\centering
\includegraphics[width=\columnwidth]{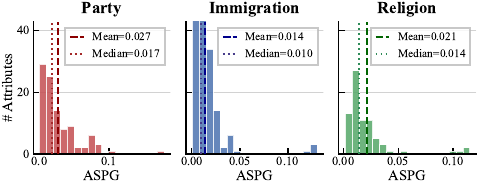}
\caption{Distribution of ASPG values across attributes for each task. Most attributes carry low predictive information individually, while a small minority contributes disproportionately large predictive gain.}
\label{fig:aspg_dist}
\end{figure}
Figure~\ref{fig:aspg_dist} shows the distribution of ASPG values across attributes for each task. In all three tasks, the distribution is strongly right-skewed: most attributes carry low predictive information individually, while a small minority contributes disproportionately large predictive gain.

\subsection{Effect of Population Coverage and Profile Completeness on Distributional Bias}
\begin{figure*}[!t]
    \centering

    \begin{subfigure}[t]{0.23\textwidth}\centering
        \includegraphics[width=\linewidth]{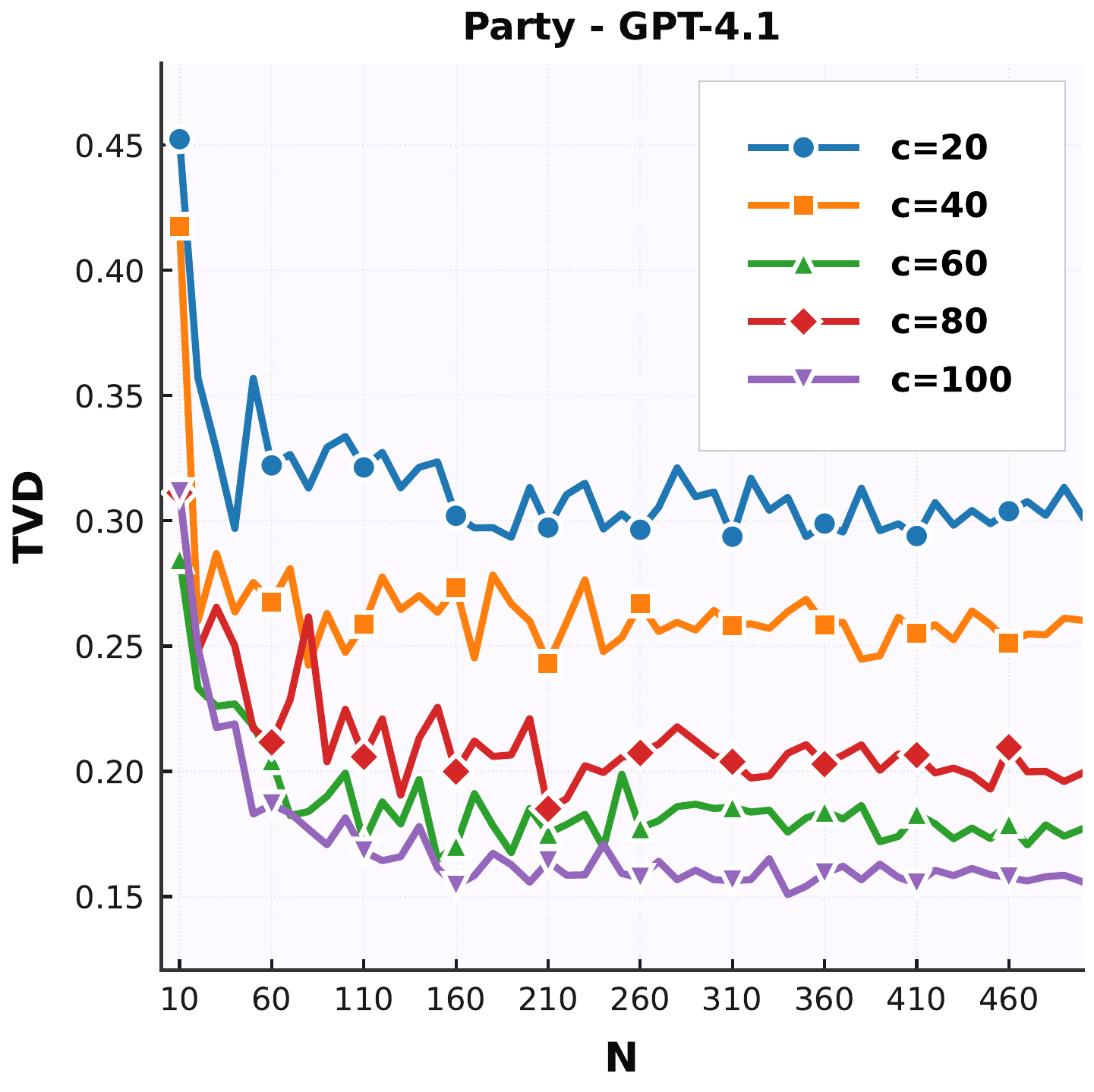}
    \end{subfigure}\hfill
    \begin{subfigure}[t]{0.23\textwidth}\centering
        \includegraphics[width=\linewidth]{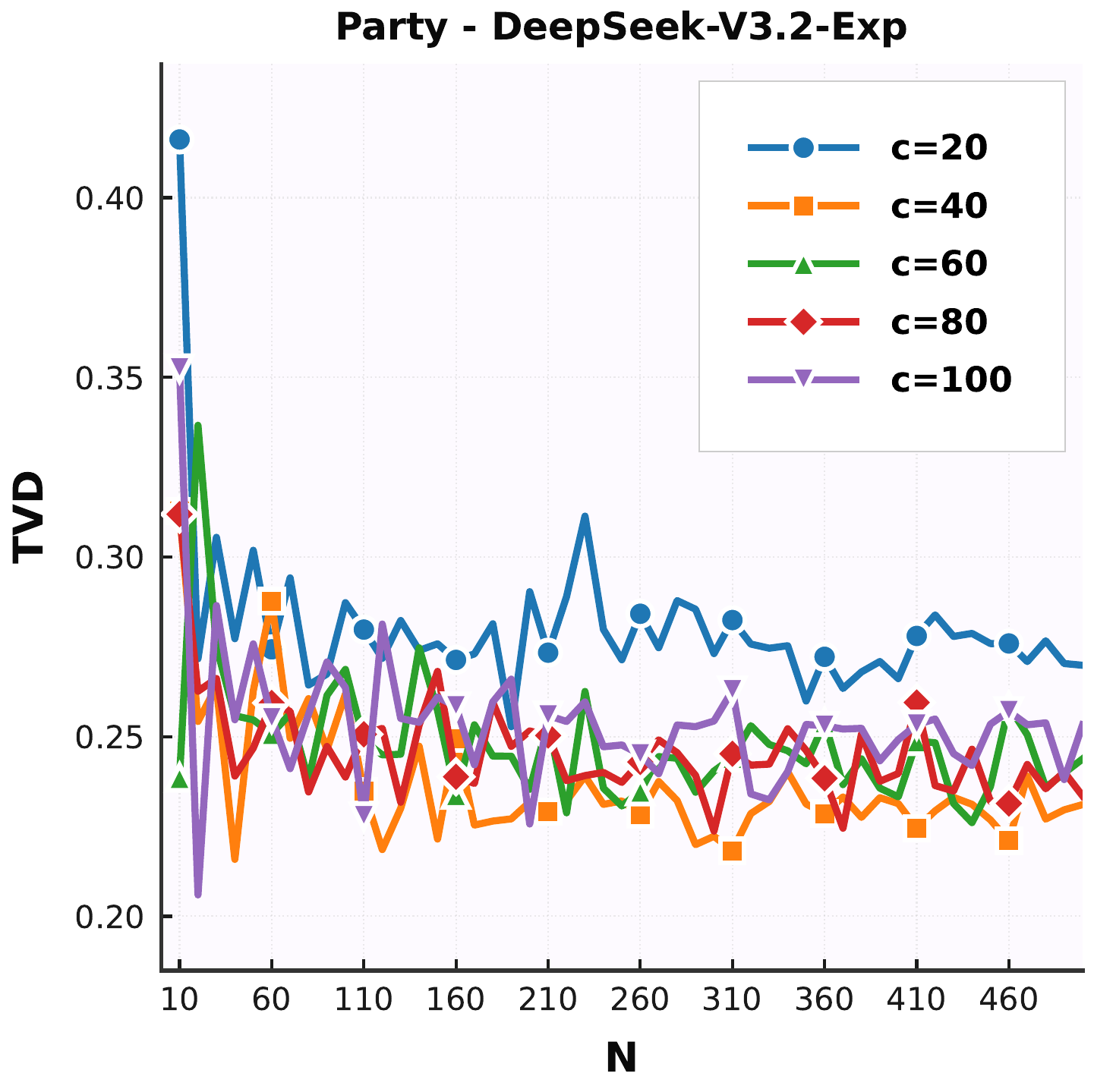}
    \end{subfigure}\hfill
    \begin{subfigure}[t]{0.23\textwidth}\centering
        \includegraphics[width=\linewidth]{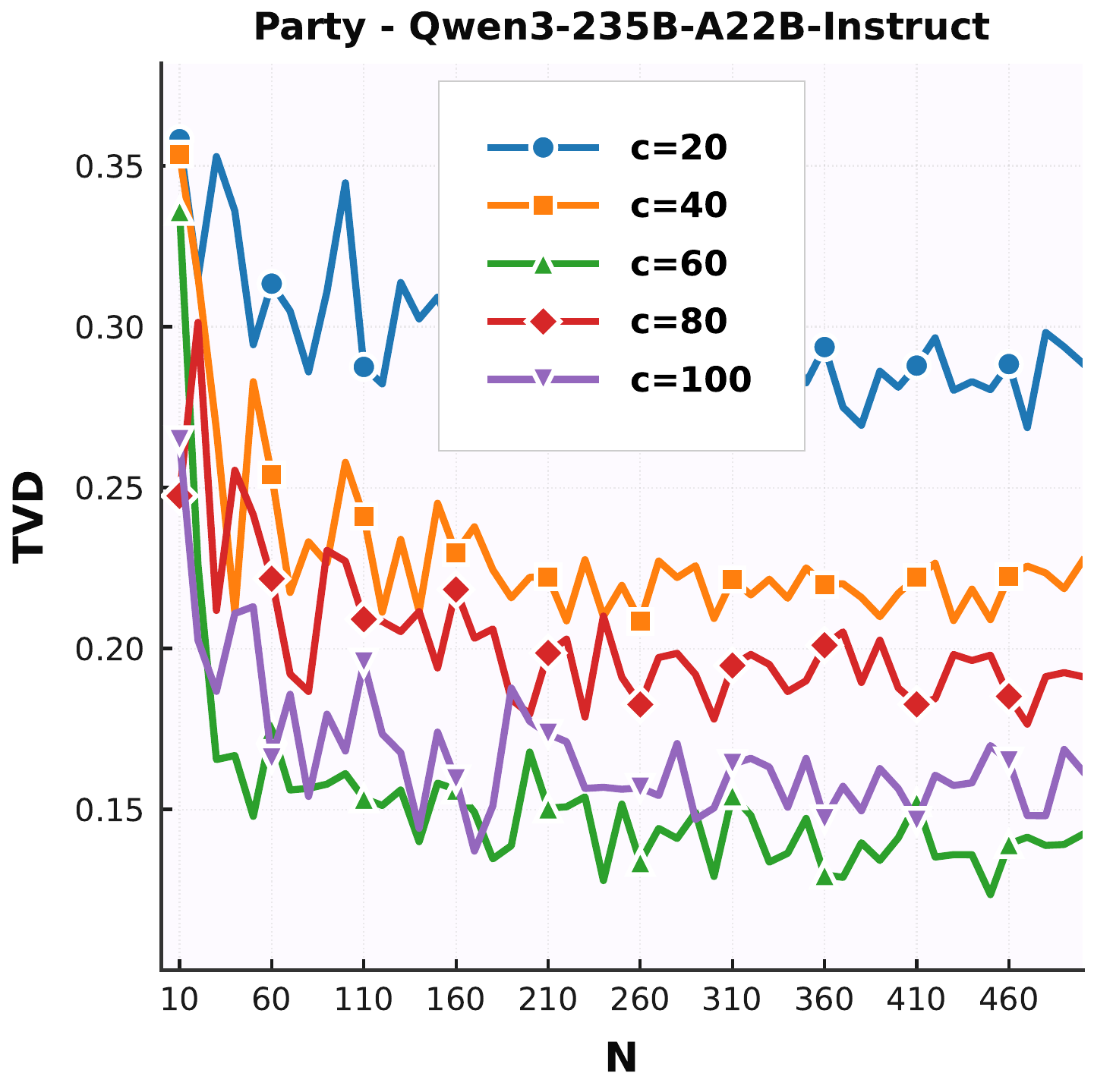}
    \end{subfigure}\hfill
    \begin{subfigure}[t]{0.23\textwidth}\centering
        \includegraphics[width=\linewidth]{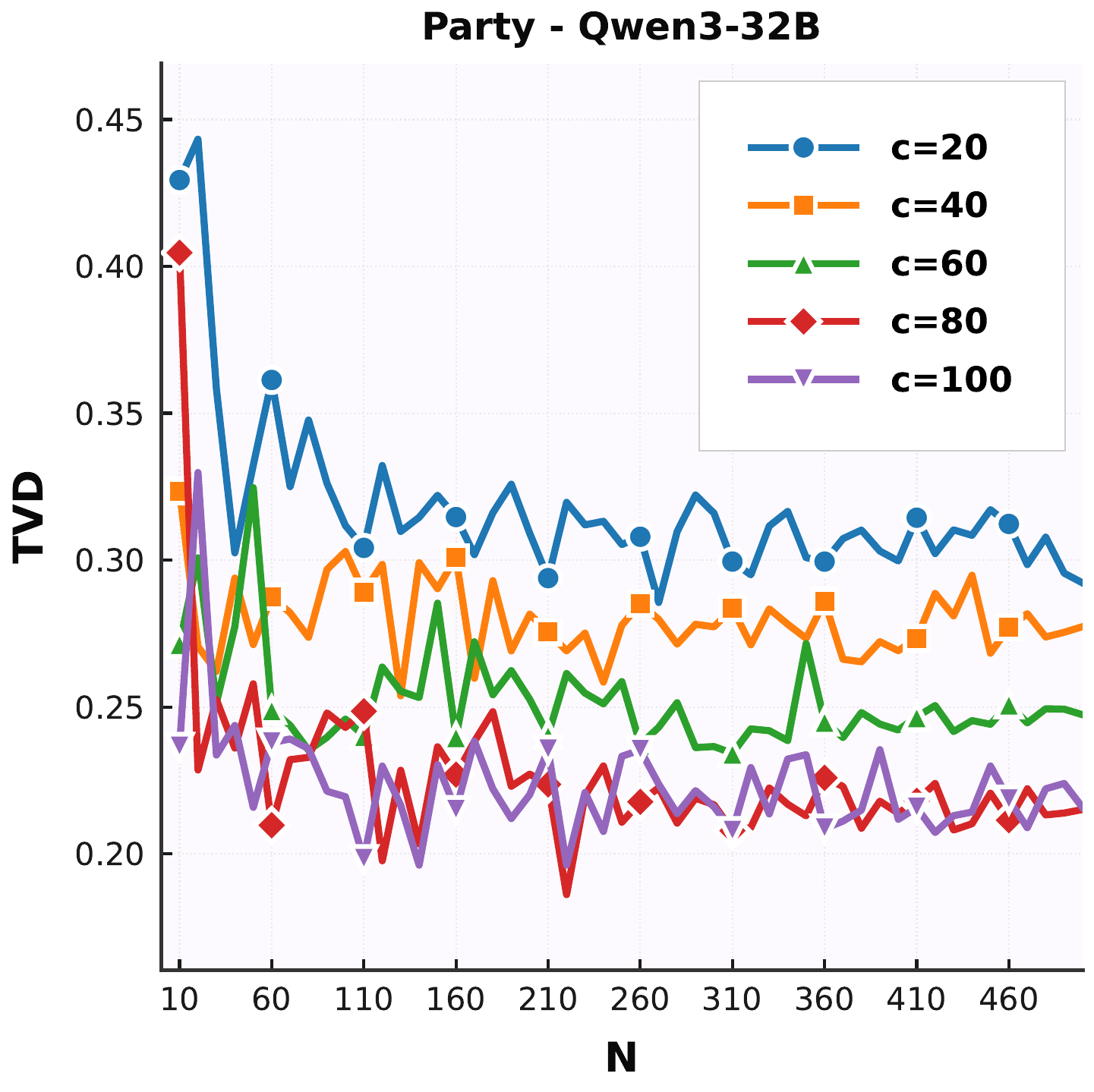}
    \end{subfigure}
    
    \vspace{2mm}
    \vspace{2mm}

    \caption{TVD Scaling with Population Coverage $N$ under Varying Profile Completeness $c$.}
    \label{app:more_exp}
\end{figure*}
To further test ReliMap's core claim, we ask whether population-level distribution errors mainly reflect
\emph{estimation noise} from limited population coverage or \emph{systematic bias} from insufficient profile information.
In the Party scenario, we vary population coverage $N$ while fixing profile completeness $c$, and measure distributional deviation by TVD.
Results are reported in Figure~\ref{app:more_exp}.

Across models, TVD decreases rapidly as $N$ increases and then quickly plateaus, indicating that larger $N$ primarily improves stability by reducing variance.
In contrast, increasing $c$ produces a more consistent downward shift of the curves, suggesting that the remaining gap is dominated by systematic bias that cannot be removed by sampling more agents alone.
Despite differences in absolute performance and smoothness across models, the trend is robust and aligns with ReliMap: $N$ mainly controls estimation variance, whereas $c$ more directly determines distributional alignment.
These results motivate future work on constructing higher-information profiles and explicitly optimizing distribution-level objectives to reduce residual bias.

\section{Alternative Metrics for R2}
\label{app:alt_metrics}

Besides TVD, we also computed KL divergence, Jensen--Shannon divergence (JSD), and Earth Mover's Distance (EMD) for R2 and checked whether the main R2 findings remain visible under these alternative metrics. Table~\ref{tab:alt_metrics_c} reports the four metrics under varying profile completeness $c$, and Table~\ref{tab:alt_metrics_n} reports them under varying population coverage $N$ (both on the Party task).

The trends are consistent across all four metrics. For profile completeness, all metrics show the same pattern: profile conditioning usually reduces population-level mismatch relative to no conditioning ($c=0$), while local non-monotonic changes persist across completeness levels, consistent with Finding 5. For population coverage, all four metrics decrease as $N$ increases, with most gains appearing at smaller $N$ before the curves stabilize, consistent with Finding 6.

\begin{table}[h]
\centering
\small
\setlength{\tabcolsep}{4pt}
\begin{tabular}{@{}llrrrr@{}}
\toprule
\textbf{Model} & $c$ & \textbf{TVD} & \textbf{KL} & \textbf{JSD} & \textbf{EMD} \\
\midrule
GPT-4.1 & 0 & 0.6704 & 10.2946 & 0.2970 & 1.5623 \\
 & 20 & 0.2982 & 1.1940 & 0.0877 & 0.5902 \\
 & 40 & 0.2571 & 0.6908 & 0.0641 & 0.5909 \\
 & 60 & 0.1751 & 0.4742 & 0.0398 & 0.4494 \\
 & 80 & 0.2008 & 0.5499 & 0.0446 & 0.3849 \\
 & 100 & 0.1551 & 0.6279 & 0.0441 & 0.3967 \\
\midrule
Qwen3-8B & 0 & 0.6891 & 6.3992 & 0.2996 & 1.5668 \\
 & 20 & 0.3443 & 0.8956 & 0.0907 & 0.7080 \\
 & 40 & 0.3396 & 0.9203 & 0.0873 & 0.7567 \\
 & 60 & 0.2934 & 0.3075 & 0.0638 & 0.7136 \\
 & 80 & 0.3752 & 0.4198 & 0.0814 & 0.9039 \\
 & 100 & 0.3148 & 0.4644 & 0.0671 & 0.4782 \\
\midrule
Qwen3-235B-A22B & 0 & 0.2706 & 3.0733 & 0.0819 & 0.7603 \\
 & 20 & 0.2866 & 0.5523 & 0.0692 & 0.6209 \\
 & 40 & 0.2189 & 0.5179 & 0.0436 & 0.4849 \\
 & 60 & 0.1320 & 0.3263 & 0.0249 & 0.3293 \\
 & 80 & 0.1881 & 0.3889 & 0.0335 & 0.3919 \\
 & 100 & 0.1537 & 0.4955 & 0.0318 & 0.3230 \\
\bottomrule
\end{tabular}
\caption{R2 metrics under varying profile completeness $c$ on the Party task. All four metrics exhibit the same qualitative trends as TVD.}
\label{tab:alt_metrics_c}
\end{table}

\begin{table}[h]
\centering
\small
\setlength{\tabcolsep}{4pt}
\begin{tabular}{@{}llrrrr@{}}
\toprule
\textbf{Model} & $N$ & \textbf{TVD} & \textbf{KL} & \textbf{JSD} & \textbf{EMD} \\
\midrule
GPT-4.1 & 10 & 0.2258 & 3.5638 & 0.0701 & 0.4924 \\
 & 50 & 0.1909 & 3.5109 & 0.0618 & 0.4939 \\
 & 100 & 0.1739 & 2.5949 & 0.0510 & 0.4184 \\
 & 150 & 0.1703 & 1.9460 & 0.0518 & 0.3688 \\
 & 200 & 0.1689 & 1.7412 & 0.0514 & 0.3752 \\
 & 250 & 0.1611 & 0.6213 & 0.0441 & 0.4321 \\
\midrule
Qwen3-8B & 10 & 0.3803 & 5.7937 & 0.1330 & 0.8403 \\
 & 50 & 0.3261 & 2.4756 & 0.0906 & 0.5418 \\
 & 100 & 0.3295 & 1.0649 & 0.0804 & 0.5107 \\
 & 150 & 0.2872 & 1.3549 & 0.0616 & 0.4396 \\
 & 200 & 0.3162 & 0.9554 & 0.0717 & 0.5025 \\
 & 250 & 0.3129 & 0.4890 & 0.0688 & 0.4532 \\
\midrule
Qwen3-235B-A22B & 10 & 0.3183 & 3.8628 & 0.0945 & 0.6335 \\
 & 50 & 0.1626 & 1.9375 & 0.0426 & 0.3724 \\
 & 100 & 0.1814 & 1.9468 & 0.0472 & 0.4196 \\
 & 150 & 0.1541 & 0.7633 & 0.0351 & 0.4352 \\
 & 200 & 0.1524 & 0.7139 & 0.0316 & 0.3590 \\
 & 250 & 0.1507 & 1.2401 & 0.0371 & 0.3503 \\
\bottomrule
\end{tabular}
\caption{R2 metrics under varying population coverage $N$ on the Party task. All four metrics decrease as $N$ increases and stabilize at larger $N$, consistent with the TVD-based results.}
\label{tab:alt_metrics_n}
\end{table}

\section{Robustness Analysis}

To examine whether our findings are sensitive to implementation-level choices, we conduct robustness tests on prompt design and decoding temperature using the Party task across three models. Each setting is evaluated over 5 independent sampling runs, with results reported as mean individual-level simulation accuracy (ACC).

\subsection{Prompt Variations}

We test two additional prompt variants alongside the original prompt described in Appendix C.

\textbf{Original prompt} follows the role-playing instruction described in Appendix C, asking the model to fully immerse itself in the respondent's persona.

\textbf{Prompt\_v1} instructs the model to predict the respondent's most probable choice instead of role-playing: \textit{``You are analyzing responses for the International Social Survey Programme. Given the demographic and attitudinal profile below, predict the most likely response this individual would provide to the survey question.''}

\textbf{Prompt\_v2} directly instructs the model to adopt the role of the respondent with a minimal instruction: \textit{``You're taking part in an international survey. Answer as a real person with these traits:''}

\subsection{Temperature}

We test decoding temperature $t = 0.6$ alongside the default greedy decoding ($t = 0$) using the original prompt template.

\begin{table*}[h]
\centering
\small
\begin{tabular}{llrrrrrrrr}
\toprule
& & & \multicolumn{6}{c}{$c$ (\%)} \\
\cmidrule(lr){4-9}
Model & Prompt & Temperature & 0 & 20 & 40 & 60 & 80 & 100 \\
\midrule
\multirow{4}{*}{Qwen3-8B}
& Original & greedy & 26.53& 26.72& 30.76& 31.96& 31.72& 28.78\\
& Prompt\_v1 & greedy & 28.40 & 31.00 & 32.57 & 34.16 & 32.08 & 33.26 \\
& Prompt\_v2 & greedy & 26.40 & 28.07 & 29.81 & 31.29 & 30.81 & 31.95 \\
& Original & $t=0.6$ & 26.52 & 27.22 & 31.05 & 32.88 & 31.87 & 32.53 \\
\midrule
\multirow{4}{*}{Qwen3-32B}
& Original & greedy & 25.76& 30.32& 33.32& 37.14& 32.26& 32.90\\
& Prompt\_v1 & greedy & 27.34 & 33.08 & 35.02 & 38.70 & 36.32 & 37.78 \\
& Prompt\_v2 & greedy & 26.38 & 30.90 & 33.26 & 35.08 & 32.91 & 33.68 \\
& Original & $t=0.6$ & 26.00 & 30.52 & 33.28 & 35.54 & 33.38 & 32.84 \\
\midrule
\multirow{4}{*}{Qwen3-235B-A22B-Instruct}
& Original & greedy & 12.52& 30.62& 36.34& 38.30& 38.88& 41.32\\
& Prompt\_v1 & greedy & 21.04 & 32.72 & 35.98 & 39.68 & 38.76 & 42.78 \\
& Prompt\_v2 & greedy & 18.78 & 29.28 & 35.20 & 37.44 & 38.40 & 39.66 \\
& Original & $t=0.6$ & 11.25 & 29.10 & 33.10 & 38.84 & 37.96 & 38.18 \\
\bottomrule
\end{tabular}
\caption{Individual-level simulation accuracy (ACC, \%) under different prompt variants and decoding temperatures on the Party task, supporting the robustness of Findings 1 and 2. Original refers to the default prompt used in the main experiments.}
\label{tab:robustness}
\end{table*}

Across all prompt variants and temperature settings, two patterns remain consistent. First, individual-level simulation accuracy improves with profile completeness but with diminishing marginal gains, confirming the robustness of Finding 1. Second, larger models achieve higher accuracy and continue to benefit more from richer profiles, confirming the robustness of Finding 2. While absolute accuracy varies across settings, the macro-level patterns identified in our main experiments remain stable, suggesting that our findings are not artifacts of particular implementation choices.

\section{Statistical Uncertainty Analysis}
\label{app:uncertainty}

To assess the statistical reliability of our experimental findings, we report the mean and standard deviation of the individual-level accuracy (R1) across 5 fixed random seeds (2025, 2026, 2027, 2028, 2029), where randomness stems from the attribute subset sampling procedure.

\begin{figure*}[h]
    \centering
    \includegraphics[width=\linewidth]{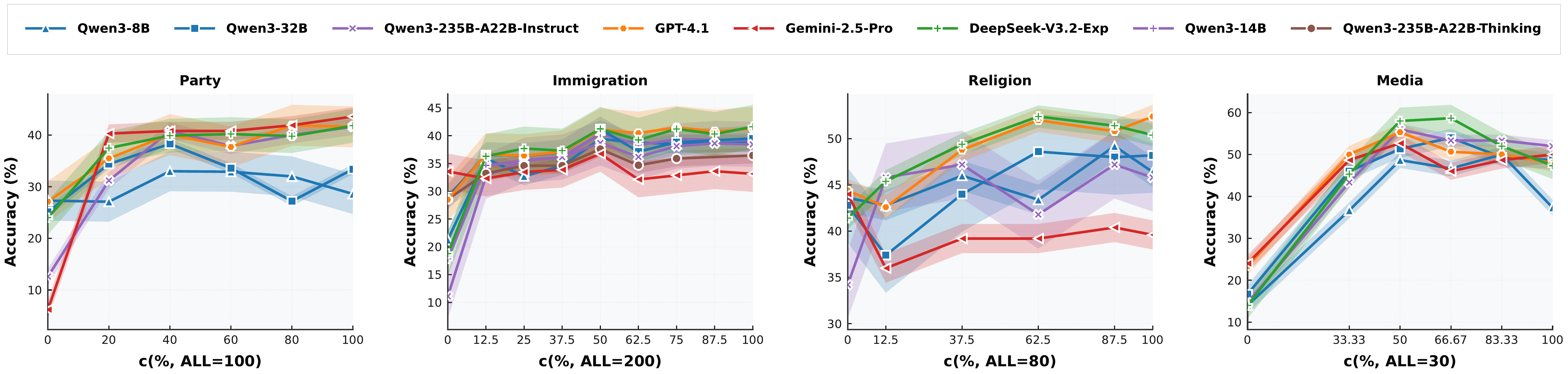}
    \caption{Individual-level accuracy (R1) as a function of profile completeness $c$ across all four tasks, with $\pm$1 standard deviation shaded regions computed over 5 fixed random seeds (2025--2029). Variance stems from attribute subset sampling. Despite variability in some tasks, the overall trends remain consistent across models.}
    \label{fig:std}
\end{figure*}

Figure~\ref{fig:std} replicates Figure~3(a) across all four tasks (Party, Immigration, Religion, and Media) with $\pm$1 standard deviation shaded regions. The shaded regions reflect the variability introduced by different random attribute subset samples rather than model stochasticity. As shown, while some tasks (e.g., Religion) exhibit larger variance due to the smaller dataset size, the main trends remain consistent across models and tasks, supporting the reliability of our conclusions.

\section{Use of AI Assistant}
In this paper, we only use ChatGPT\footnote{\url{https://chatgpt.com/}} for grammar proofreading and spell checking.

\end{document}